\pdfoutput=1
\PassOptionsToPackage{table,xcdraw}{xcolor}

\documentclass[11pt]{article}

\usepackage[preprint]{acl}

\usepackage{times}
\usepackage{latexsym}

\usepackage[T1]{fontenc}
\usepackage[utf8]{inputenc}

\usepackage{microtype}

\usepackage{inconsolata}

\usepackage{graphicx}
\usepackage{tabularx}
\usepackage{array}
\usepackage{xcolor}
\usepackage{booktabs}
\usepackage{amsmath}
\usepackage{amssymb}
\usepackage{subcaption}
\usepackage{arydshln}
\usepackage{newunicodechar}
\usepackage{makecell}
\usepackage{multirow}
\usepackage{float}
\usepackage{fontawesome}

\newunicodechar{✗}{$\times$}
\newunicodechar{✓}{$\checkmark$}

\usepackage[toc,page,header]{appendix} 
\usepackage{minitoc}  
\usepackage{bm}

\newcommand{\TP}{$\textcolor{red}{\rm TP\downarrow}$}
\newcommand{\TN}{$\textcolor{blue}{\rm TN\downarrow}$}
\newcommand{\AN}{$\textcolor{orange}{\rm AN\downarrow}$}
\newcommand{\AP}{$\textcolor{teal}{\rm AP\downarrow}$}
\newcommand{\TPbroad}{$\textcolor{red}{\rm TP}$}
\newcommand{\TNbroad}{$\textcolor{blue}{\rm TN}$}
\newcommand{\ANbroad}{$\textcolor{orange}{\rm AN}$}
\newcommand{\APbroad}{$\textcolor{teal}{\rm AP}$}

\newcommand{\paragraphx}[1]{\smallskip\smallskip\smallskip\noindent\textbf{#1}}

\title{How Does DPO Reduce Toxicity? 
A Mechanistic Neuron-Level Analysis
}
\author{
  \textbf{Yushi Yang\textsuperscript{1}\thanks{Correspondence to: \href{mailto:yushi.yang@oii.ox.ac.uk}{yushi.yang@oii.ox.ac.uk}}} \quad
  \textbf{Filip Sondej\textsuperscript{2}} \quad
  \textbf{Harry Mayne\textsuperscript{1}} \quad
  \textbf{Andrew Lee\textsuperscript{3}\textsuperscript{†}} \quad
  \textbf{Adam Mahdi\textsuperscript{1}}\thanks{Equal contribution.} \\
  \textsuperscript{1}University of Oxford \quad
  \textsuperscript{2}Jagiellonian University \quad
  \textsuperscript{3}Harvard University
}

\begin{document}

\doparttoc % Tell to minitoc to generate a toc for the parts
\faketableofcontents % Run a fake tableofcontents command for the partocs

\maketitle

\begin{abstract}
Safety fine-tuning algorithms reduce harmful outputs in language models, yet their mechanisms remain under-explored.
Direct Preference Optimization (DPO) is a popular choice of algorithm, but prior explanations, attributing its effects solely to dampened \textit{toxic neurons} in the MLP layers, are incomplete.
In this study, we analyse four language models (Llama-3.1-8B, Gemma-2-2B, Mistral-7B, GPT-2-Medium) 
and show that toxic neurons only account for 2.5\% to 24\% of DPO's effects across models. 
% ’s weight shifts have a more nuanced effect, 
Instead, DPO 
balances distributed activation shifts across 
all MLP neurons to create a net toxicity reduction. 
We attribute this reduction to four 
neuron groups, two aligned with reducing toxicity and two promoting anti-toxicity, whose 
combined effects replicate DPO across models. 
% We find that 
% their post-DPO activation changes depend on their 
% orientations relative to the toxicity representation. 
To further validate this understanding, we develop an activation editing method mimicking DPO through distributed shifts along a toxicity representation.  
This method outperforms DPO in reducing toxicity while preserving perplexity, 
without requiring any weight updates. 
Our work provides a mechanistic understanding of DPO and introduces an efficient, tuning-free alternative for safety fine-tuning. 
Our code is available on \href{https://github.com/Yushi-Y/dpo-toxic-neurons}{\faGithub\ dpo-toxic-neurons}. 

%\href{https://github.com/Yushi-Y/dpo-toxic-neurons}{github.com/Yushi-Y/dpo-toxic-neurons}. 

\end{abstract}
% {github.com/Yushi-Y/dpo-toxic-neurons}.

% in the anonymous repository: 
% \href{https://anonymous.4open.science/r/dpo-mlp-toxic}
% {anonymous.4open.science/r/dpo-mlp-toxic}.

\section{Introduction}
% wide shot
The growing capabilities of large language models (LLMs) also lead to the encoding of undesirable behaviours \cite{gehman2020RealToxicityPromptsevaluatingneuraltoxic, gallegos2024biasfairnesslargelanguage}. 
To mitigate harmful outputs, researchers have developed fine-tuning algorithms to  prioritise human-preferred responses through reward modelling~\cite{schulman2017proximalpolicyoptimizationalgorithms, shao2024deepseekmathpushinglimitsmathematical}. 
Among these, Direct Preference Optimization (DPO) has been a popular algorithm given its simplicity to directly optimise the policy model \cite{rafailov2024directpreferenceoptimizationlanguage}.
While these algorithms effectively reduce harmful behaviours at the output level, there is limited mechanistic understanding of how they achieve this internally. 
This gap limits our ability to explain their vulnerability to jailbreaks and adversarial fine-tuning \cite{wei2023jailbrokendoesllmsafety, yang2023shadowalignmenteasesubverting, qi2023finetuningalignedlanguagemodels}.
% shayegani2023surveyvulnerabilitieslargelanguage, geiping2024coercingllmsrevealalmost, 

% niche
Recent studies found that fine-tuning algorithms lead to superficial changes, allowing models to retain the undesirable capabilities \cite{jain2024makesbreakssafetyfinetuning, yang2023shadowalignmenteasesubverting}.
% This may explain why they are vulnerable to weight-based attacks such as weight perturbations and adversarial fine-tuning \cite{wei2023jailbrokendoesllmsafety, yang2023shadowalignmenteasesubverting, qi2023finetuningalignedlanguagemodels}.
In particular, \citet{lee2024mechanisticunderstandingalignmentalgorithms} suggested that DPO reduces toxicity by dampening the activations of a few \textit{toxic neurons} in the MLP layers.
% gap
While this offers an intuitive explanation, it assumes that toxicity is localised to a small subset of neurons. However, this is a strong claim that may oversimplify how safety fine-tuning works. 
% It leaves open a key question: does this account fully explain how DPO reduces toxic outputs? 
% , or are other mechanisms involved
% solution

In this paper, we show that this explanation is incomplete, and offer a more comprehensive analysis of DPO's mechanism across four LLMs: Llama-3.1-8B, Gemma-2-2B, Mistral-7B and GPT-2-Medium.

\paragraphx{Toxic neurons are not enough to explain DPO.}
We use activation patching to isolate the role of toxic neurons, and observe only a partial drop in toxicity across models (2.5\% to 24\%) compared to DPO.
Where, then, does the rest of DPO’s toxicity reduction come from?

\paragraphx{Four neuron groups reduce toxicity.}
We show that DPO induces more nuanced, distributed activation shifts across all MLP neurons than previously suggested. 
% resulting in a net reduction in toxicity. 
% just bypassing toxic neurons 
We identify four mutually exclusive neuron groups that consistently contribute to the toxicity reduction across models.
We find that their post-DPO activation changes depend on their orientation relative to the toxicity representation. 
Using activation patching, we show that their combined influence can match or even exceed the toxicity reduction achieved by DPO.

\paragraphx{Activation editing to replicate DPO.}
To validate our understanding, 
we develop a simple activation editing method to replicate DPO.
Unlike the previous post-hoc patching analyses, our method does not rely on access to post-DPO activations, nor does it require weight updates or pairwise preference data. 
% pairwise preference data are pairs of text sharing the same prompt/context, not classification data
% small, distributed 
% solely
Instead, we leverage our observations to edit activations based on the orientation of MLP weights relative to a toxicity representation. 
This method consistently outperforms DPO across models, showing 
that DPO-like effects can be achieved with minimal intervention and without fine-tuning. 
% solely

\section{Related Work} 
\label{sec:related_work}
Here we review the DPO algorithm, the Transformer MLP layers and related work on mechanisms of safety fine-tuning algorithms.

\paragraphx{DPO algorithm.}
DPO is a fine-tuning algorithm designed to align LLMs with pairwise human preference data \cite{rafailov2024directpreferenceoptimizationlanguage}.
Given $N\in\mathbb{Z}$ preference data triplets 
\[
\left\{ \left( x^{(i)}, y_+^{(i)}, y_-^{(i)} \right) \right\}_{i=1}^{N},
\]  
where $x$ is the input prompt, and $y_+, y_-$ are pairwise preferred and non-preferred continuations, 
DPO fine-tunes a policy model $\pi_{\theta}(y_+\! \mid\! x)$ that assigns a higher likelihood to $y_+^{(i)}$ compared to $y_-^{(i)}$. 

The DPO loss is defined as
\[
\mathcal{L}_{\text{DPO}}(\theta) = -\log\sigma\!\left( \beta \left( r_{\theta}(x, y^+) - r_{\theta}(x, y^-) \right) \right),
\] 
where $\sigma$ is the sigmoid function, $\beta$ is a temperature hyperparameter and $r_\theta$ is the derived reward regularised using the reference model $\pi_{\text{ref}}$, that is
\begin{equation*}
    r_{\theta}(x, y) = \log \frac{\pi_{\theta}(y \mid x)}{\pi_{\text{ref}}(y \mid x)}.
\end{equation*}

\paragraphx{MLP layers.}\label{sec:mechanism_mlp}
MLPs apply two linear transformations with a non-linearity $\sigma$ in between:
\begin{equation*}
\text{MLP}^{\ell}(\mathbf{x}^{\ell}) = \sigma\left(W_K^{\ell}\mathbf{x}^{\ell}\right) W_V^{\ell},
\end{equation*}
where $W_K^{\ell}, W_V^{\ell} \in \mathbb{R}^{d_{\text{mlp}} \times d}$, $d_\text{mlp}$ and $d$ are the dimensions of MLP hidden layers and the residual stream.
MLPs can be re-expressed as:
\begin{equation}
\label{eq:mlp_exp}
\text{MLP}^{\ell}(\mathbf{x}^{\ell}) = \sum_{i=1}^{d_{\text{mlp}}} m_i^{\ell} \mathbf{v}_i^{\ell}, \quad m_i^{\ell} = \sigma(\mathbf{k}_i^{\ell} \cdot \mathbf{x}^{\ell}),
\end{equation}
% \begin{align*}
% \label{eq:mlp_exp}
% \text{MLP}^{\ell}(\mathbf{x}^{\ell}) = \sum_{i=1}^{d_{\text{mlp}}} m_i^{\ell} \mathbf{v}_i^{\ell}, \\
%     m_i^{\ell} = \sigma(\mathbf{k}_i^{\ell} \cdot \mathbf{x}^{\ell})
% \end{align*}
where $\mathbf{k}_i^{\ell}, \mathbf{v}_i^{\ell} \in \mathbb{R}^{d}$ are the $i$-th row of $W_K^{\ell}$ and $W_V^{\ell}$, respectively.
For each MLP neuron $i$, 
we refer to $\mathbf{v}_i^\ell$ as its \textit{value vector} following \citet{geva2022transformerfeedforwardlayersbuild} and \citet{lee2024mechanisticunderstandingalignmentalgorithms}. 
The scalar 
\( m_i^{\ell} \in \mathbb{R} \) is an \textit{activation score} controlling the scaling of the value vector \( \mathbf{v}_i^\ell \). 
This means an MLP layer writes to the residual stream $d_{\text{mlp}}$ times, once per neuron, via the activation-weighted value vector $m_i^{\ell} \mathbf{v}_i^{\ell}$. 

Recent models (Llama, Gemma, Mistral) replace MLPs with Gated Linear Units (GLUs) \cite{shazeer2020gluvariantsimprovetransformer}.
GLUs can similarly be expressed as a weighted sum of value vectors as in \eqref{eq:mlp_exp}, where each weight is determined by some non-linear activation.
See Appendix~\ref{appendix:glu} for details.

\paragraphx{Mechanisms of safety fine-tuning algorithms.} 
Recent studies have shown that fine-tuning induces superficial weight changes, leaving most pre-trained capabilities intact.
\citet{jain2023mechanisticallyanalyzingeffectsfinetuning} found that fine-tuning on synthetic tasks produces `wrappers', i.e. localised weight changes in later layers optimised for each task.
% \citet{jain2024makesbreakssafetyfinetuning} found that safety fine-tuning projects unsafe inputs into the null space of models through minimal weight changes. 
\citet{qi2024safetyalignmentjusttokens} found that aligned models primarily adapt their generative distribution in the first few output tokens. 
\citet{wei2024assessingbrittlenesssafetyalignment} showed that pruning just 3\% of targeted parameters can undo safety alignment, highlighting the brittleness of safety mechanisms. 
These findings suggest that safety fine-tuning reduces harmful outputs through subtle, targeted weight changes rather than large-scale rewiring. 

\citet{lee2024mechanisticunderstandingalignmentalgorithms} studied the mechanisms of how DPO reduces toxic outputs, attributing its effects to dampened activations of a few toxic MLP value vectors. 
We revisit this claim and find it to be incomplete, as shown in Section~\ref{sec:dpo_incomplete}.

\section{Experimental Setup}
\label{sec:experiment_setup}
Here we describe the methods used in this study, including the data and models, linear probes, projections and activation patching.
\subsection{Data and Models}
\label{subsec:data_and_mdoels}

\paragraphx{Toxicity-eliciting prompts.} 
We use the `challenge' subset (N=1,199) of \textit{RealToxicityPrompts} \cite{gehman2020RealToxicityPromptsevaluatingneuraltoxic} to elicit toxic outputs from each model.
This subset is designed to trigger extremely toxic completions, making it a strong testbed for safety fine-tuning algorithms.

\paragraphx{Models.} 
We study four pre-trained LLMs: Llama-3.1-8B \cite{grattafiori2024llama3herdmodels}, Gemma-2-2B \cite{gemmateam2024gemma2improvingopen}, Mistral-7B \cite{jiang2023mistral7b} and GPT-2 Medium \cite{radford2019language}.
GPT-2 Medium is included to compare with claims made in \citet{lee2024mechanisticunderstandingalignmentalgorithms}.
We generate toxic outputs from each LLM using greedy decoding.
Appendix~\ref{appendix:mlp_spec} provides the MLP specification for each model.

\paragraphx{Evaluation metrics.} 
We report three metrics: \textit{toxicity scores} using Detoxify \cite{Hanu_Detoxify_2020}, 
a BERT model fine-tuned for toxicity classification that assigns a likelihood score for a text being toxic; 
% (higher scores indicate greater toxicity)
\textit{log perplexity}, the average negative log-likelihood of generated tokens on the Wikitext-2 dataset \cite{merity2016pointersentinelmixturemodels}; \textit{F1 scores}, the harmonic mean of precision and recall based on token overlap on 2,000 Wikipedia sentences \cite{lee2024mechanisticunderstandingalignmentalgorithms}. 
The latter two metrics measure general language quality, where F1 complements perplexity by capturing exact token matches.

\paragraphx{DPO training.} 
We implement DPO using 24,576 toxicity contrastive pairs generated from Wikitext-2 prompts \cite{lee2024mechanisticunderstandingalignmentalgorithms}. 
See Appendix~\ref{appendix:dpo_hyperparameters} for the training hyperparameters.

\subsection{Per-Neuron Toxicity Contributions}
\label{subsec:per_neuron_toxicity_contributions}
We measure per-neuron contributions to toxicity by projecting activations onto linear toxicity probes. 
We describe how we extract the probes, validate their effects and compute per-neuron contributions.

\paragraphx{Linear probes.}
To extract toxicity representations, 
we train linear probes $W_\text{Toxic}$ to classify toxic versus non-toxic inputs for each model. 
The probe is trained on the final-layer residual stream  ${\bar{\mathbf{x}}^{L-1}}$, averaged across all token positions:
\begin{equation*}
P(\text{toxic} \mid{\bar{\mathbf{x}}^{L-1}}) = \sigma(W_{\text{Toxic}} {\bar{\mathbf{x}}^{L-1}} + b),
\end{equation*}
where $\sigma$ is the sigmoid function, \( W_{\text{Toxic}} \in \mathbb{R}^{d} \) is the learned probe vector.  
%\( b \in \mathbb{R} \) is the bias term.
We use the \textit{Jigsaw Toxic Comment Classification} dataset \cite{jigsaw-toxic-comment-classification-challenge}, which contains 561,808 comments labelled as toxic or non-toxic.

Across the four models, all  linear probes achieve over 91\% test accuracy using a 90:10 train/test split (Appendix Table~\ref{tab:validation_accuracy}).
When projected onto each model's vocabulary space via the unembedding matrix, i.e. through LogitLens \cite{logit_lens}, the trained probes predominantly map to toxic tokens (Table~\ref{tab:model_top_tokens}).

\begin{table}[ht]
\centering
\renewcommand{\arraystretch}{1.2}
\setlength{\tabcolsep}{2pt}
\arrayrulecolor{black}
% The top tokens for four toxic probes are obtained using the Logit Lens. 
\caption{\textit{The four toxicity probes predominantly project to toxic tokens in the vocabulary space.}
\textcolor{red}{Warning: these examples are highly offensive.}}
\label{tab:logit_lens}
\small
% \vspace{-0.5em}
 \begin{tabular}{@{}l p{5.8cm}@{}}
        \toprule
        \textbf{Model} & \textbf{Top tokens projected by probes} \\ 
        \midrule
        GPT-2-355M & f*ck, c*nt, a**hole, holes, d*ck, wh*re\\ 
        % Godd, wh*re 
        Llama-3.1-8B & en, kommen, F*CK, iyah, f*ck, dirty\\
        Gemma-2-2B & rungsseite, fu*k, Fu*king, SH*T, a**hole \\ % a**holes
        % Gemma-2-9B & plufieurs, fu*k, fu*ker, fu*king, Fu*k, find \\ 
        Mistral-7B & sh*t, f*ck,  assh, bullsh*t, f*cked, a**hole\\ % , upid, dick, shitty
        \bottomrule
    \end{tabular}
\label{tab:model_top_tokens}
\end{table}

\begin{table*}[t!]
\renewcommand{\arraystretch}{1.4}  
\setlength{\tabcolsep}{2pt} 
\centering
\small  
\caption{\textit{Toxicity (Toxic), log perplexity (PPL), and F1 scores with activation patching and editing.}
Across models, patching toxic neurons—whether those with toxic tokens or the top 256—yields only a limited drop in toxicity scores than DPO (Section~\ref{sec:dpo_incomplete}).
In contrast, patching all four of our identified neuron groups matches or outperforms DPO (Section~\ref{subsec:neuron_groups}).
Our activation editing method can outperform DPO, steering with probe and patching all four groups (Section~\ref{subsec:replicating_dpo_with_activation_editing}).
\colorbox{green!20}{Green} shows the editing parameters that best compete with DPO while preserving F1 scores.
%  and perplexity
}
% This method requires only a toxicity probe to adjust activations, without relying on DPO, and minimally affects perplexity while incurring only a slight F1 drop.
\label{tab:all_patching_results_main}
\begin{tabular}{l|l|p{0.82cm}p{0.73cm}p{0.8cm}|p{0.82cm}p{0.73cm}p{0.8cm}|p{0.82cm}p{0.73cm}p{0.8cm}|p{0.82cm}p{0.73cm}p{0.8cm}}
\toprule
\quad\textbf{Type} &\qquad \textbf{Intervention} & \multicolumn{3}{c|}{\textbf{GPT-2-355M}} & \multicolumn{3}{c|}{\textbf{Llama-3.1-8B}} & \multicolumn{3}{c|}{\textbf{Gemma-2-2B}} & \multicolumn{3}{c}{\textbf{Mistral-7B}} \\  
%\cline{3-14}
& & \textbf{Toxic} & \textbf{PPL} &\,\,\, \textbf{F1} & \textbf{Toxic} & \textbf{PPL} &\,\,\, \textbf{F1} & \textbf{Toxic} & \textbf{PPL} & \,\,\,\textbf{F1} & \textbf{Toxic} & \textbf{PPL} & \,\,\,\textbf{F1} \\
\hline
\multirow{3}{*}{Baselines} 
& None      & 0.545 & 3.08 & 0.193 & 0.496 & 1.94 & 0.225 & 0.488 & 4.61 & 0.231 & 0.507 & 1.76 & 0.221 \\
% & Forbid toxic tokens & 
% 0.541 & 3.07 & 0.193 &  
% 0.492 & 1.94 & 0.225 & 
% 0.484 & 4.62 & 0.231 & 
% 0.451 & 1.76 & 0.232 \\
& Steering with probe  & 0.310 & 3.19 & 0.191 & 0.335 & 2.72 & 0.187 & 0.260 & 5.52 & 0.228 & 0.350 & 2.23 & 0.220 \\
& DPO       & 0.210 & 3.15 & 0.195 & 0.241 & 2.69 & 0.221 & 0.245 & 5.15 & 0.228 & 0.191 & 2.01 & 0.223 \\
\hline
\multirow{6}{*}{\makecell{Activation\\patching\\(Sec~\ref{subsec:neuron_groups})}}
& Patch toxic neurons & 
0.479 & 3.09 & 0.193 &  
0.491 & 1.94 & 0.225 & 
0.487 & 4.61 & 0.231 &
0.505 & 1.76 & 0.232\\
& Patch 256 neurons & 
0.465 & 3.07 & 0.193 & 
0.488 & 1.94 & 0.225 & 
0.482 & 4.61 & 0.231 & 
0.455 & 1.76 & 0.232 \\ 
%\cline{2-14}
& Patch \textcolor{red}{$\rm TP$↓} & 0.407 & 3.07 & 0.191 & 0.488 & 1.94 & 0.223 & 0.470 & 4.87 & 0.235 & 0.502 & 1.80 & 0.229 \\
& Patch \textcolor{red}{$\rm TP$↓}+\textcolor{orange}{$\rm AN$↓} & 0.216 & 3.08 & 0.183 & 0.465 & 1.94 & 0.221 & 0.337 & 4.59 & 0.224 & 0.307 & 1.76 & 0.227 \\
& 
Patch \textcolor{red}{$\rm TP$↓}+\textcolor{orange}{$\rm AN$↓}+\textcolor{blue}{$\rm TN$↓} 
% Patch three groups
& 0.194 & 3.08 & 0.170 & 0.391 & 1.94 & 0.208 & 0.307 & 4.59 & 0.217 & 0.238 & 1.81 & 0.218 \\
& Patch four groups & 0.139 & 3.08 & 0.170 & 0.278 & 1.94 & 0.207 & 0.260 & 4.58 & 0.213 & 0.138 & 1.78 & 0.209 \\
% \cline{2-14}
% & Patch all ↑ neurons & 0.853 & \textcolor{red}{6.05} & 0.154 & 0.536 & 2.64 & 0.184 & 0.686 & 4.58 & 0.199 & 0.611 & 1.78 & 0.199 \\
\hline
\multirow{3}{*}{\makecell{Activation \\editing\\(Sec~\ref{subsec:replicating_dpo_with_activation_editing},  
\\probe-based)}} 
% \\
% (by descending
% \\cossim)
% by abs cossim (decending)
& {$\alpha=0.01, \beta=0.8$} & 
0.123 & 3.08 & 0.179 & 
0.045 & 2.19 & 0.186 & 
0.199 & 4.54 & 0.188 & 
0.038 & 1.77 & 0.179 \\ 
& $\alpha=0.01$, $\beta=0.6$ & 
0.159 & 3.08 & 0.181 & 
0.183 & 2.11 & 0.193 & 
0.200 & 4.56 & 0.201 & 
0.098 & 1.77 & 0.196 \\ 
& {$\mathbf{\alpha=0.01, \beta=0.55}$} & 
\cellcolor{green!20}
0.203 & 3.08 & 0.183 & 
\cellcolor{green!20}
0.241 & 1.96 & 0.196 & 
\cellcolor{green!20}
0.216 & 4.56 & 0.210 & 
\cellcolor{green!20}
0.125 & 1.77 & 0.202 \\ 
%& {$\alpha=0.05, \beta=0.5$} & 
%0.211 & 3.08 & 0.184 & 
%0.299 & 1.96 & 0.200 & 
%0.260 & 4.56 & 0.204 & 
%0.264 & 1.77 & 0.197\\
\hline
\multirow{3}{*}{\makecell{Activation \\editing
\\
(Sec~\ref{subsec:replicating_dpo_with_activation_editing}, 
\\probe-free)}} 
% by descending cossim
& {$\alpha=0.01, \beta=0.8$} & 
0.139 & 3.08 & 0.176 &
0.116 & 5.82 & 0.200 &
0.218 & 4.54 & 0.180 &
0.057 & 1.77 & 0.191\\ 
& {$\mathbf{\alpha=0.01, \beta=0.6}$} &
\cellcolor{green!20}0.238 & 3.08 & 0.178&
\cellcolor{green!20}0.258 & 2.28 & 0.210 &
\cellcolor{green!20}0.216 & 4.57 & 0.203 &
\cellcolor{green!20}0.162 & 1.77 & 0.200 \\ 
& {$\alpha=0.01, \beta=0.55$} & 
0.282 & 3.08 & 0.180 &
0.318 & 2.24 & 0.204 &
0.250 & 4.58 & 0.198 &
0.239 & 1.77 & 0.201 \\ 
%& {$\alpha=0.05, \beta=0.5$} & 
%0.275 & 3.08 & 0.180&
%0.322 & 2.21 & 0.201 &
%0.271 & 4.58 & 0.193&
%0.244 & 1.77 & 0.199 \\ 
\bottomrule
\end{tabular}
\end{table*}

\paragraphx{Validating linear probes.}
To further validate these probes represent toxicity, we apply \textit{activation steering} \cite{zou2025representationengineeringtopdownapproach,panickssery2024steeringLlama2contrastive} by subtracting a scaled probe \( W_{\text{Toxic}} \) from the final-layer residual stream \( \mathbf{x}^{L-1} \) at each token position:
% during inference
\begin{equation*}
\mathbf{x}^{L-1}_{\text{steered}} = \mathbf{x}^{L-1} - \alpha W_{\text{Toxic}},
\end{equation*}
where $\alpha$ is selected to preserve language quality (perplexity and F1) of pre-trained models (see Appendix Table~\ref{tab:validation_accuracy}).
Increasing $\alpha$ further reduces toxicity scores but raises perplexity (see Appendix Table~\ref{tab:subtraction_more_alpha}).
Table~\ref{tab:all_patching_results_main} shows that steering with probe consistently reduces toxicity scores across models, validating their effects in eliciting toxic outputs.
We therefore include it as a baseline for toxicity reduction.
% probe-based steering

\paragraphx{Per-neuron toxicity change via projection.}
To compute per-neuron contributions, 
we track how the toxic representation changes at each MLP neuron during DPO via its \textit{change in projection} onto the probe:
\begin{equation}
\label{eq:toxicity_reduction}
\Delta_{\text{Toxic}, i} \!= 
(m_i^\text{pre} \mathbf{v}_i^\text{pre}\!- m_i^\text{dpo} \mathbf{v}_i^\text{dpo}) \cdot \frac{W_\text{Toxic}}{\|W_\text{Toxic}\|_2},
\end{equation}
where \( m_{i}^\text{pre}\mathbf{v}_{i}^\text{pre} \) and \( m_{i}^\text{dpo}\mathbf{v}_{i}^\text{dpo} \) 
are the activated components of the \( i \)-th value vector before and after DPO; 
the activation scores \( m_{i}^\text{pre} \) and \( m_{i}^\text{dpo} \) are averaged over 20 generated tokens for all prompts in RealToxicityPrompts. 
This approach, known as \textit{direct feature attribution} \cite{makelov2024principledevaluationssparseautoencoders, arditi2024refusallanguagemodelsmediated}, 
measures how much each neuron contributes to the toxicity representation.

\subsection{Activation Patching}
\label{subsec:activation_patching_setup}
Throughout our work, we apply \textit{activation patching} \cite{zhang2024bestpracticesactivationpatching} in a counterfactual manner to isolate the effect of specific neurons on toxicity scores. 
Namely, for a pre-trained model and a set of MLP value vectors, we set their activations to match its post-DPO counterpart, based on the mean activation of 1,199 RealToxicityPrompts and 20 generated tokens per prompt. 
We then measure the resulting change in the toxicity scores.

% \begin{table}[ht]
% \centering
% \small
% \renewcommand{\arraystretch}{1.3}
% \caption{\textit{Frequent toxic tokens blocked from being generated by the models in the simple baseline.}}
% \label{tab:forbidden_toxic_tokens}
% \begin{tabular}{lllll}
% \toprule
% \texttt{fu*k} & \texttt{fu*ked} & \texttt{fu*king} & \texttt{sh*t} & 
% \texttt{sh*tty} \\
% \texttt{cr*p} & \texttt{d*mn} & \texttt{c*nt} & \texttt{a**hole} & \texttt{bullsh*t} \\
% \bottomrule
% \end{tabular}
% \end{table}

\section{Toxic Neurons Are Not Enough}
% Don't Fully Explain DPO
\label{sec:dpo_incomplete}
We start by revisiting the claims in \citet{lee2024mechanisticunderstandingalignmentalgorithms}: 
(a) DPO reduces toxicity primarily by dampening the activation of toxic neurons, 
(b) this arises from shifts in earlier layer weights.
We show here that (a) only partially explains the drop in toxicity, 
and in Section~\ref{sec:deeper_look_at_weight_shifts}, we show that the weight shifts (b) are more nuanced than simply bypassing toxic neurons.

We measure the effect of dampening toxic neurons.
We define toxic neurons by adapting the method of \citet{lee2024mechanisticunderstandingalignmentalgorithms}: we identify the top N (= 256)\footnote{This number is based on \citet{lee2024mechanisticunderstandingalignmentalgorithms}'s number (128). We double the number of accommodate larger model sizes, but see similar results with the original 128 vectors.} MLP value vectors with the highest cosine similarity to the toxic probe \( W_{\text{Toxic}} \).
In a second variant, we identify a smaller subset of interpretable value vectors.
To do so, we unembed each value vector and consider it as toxic if any of its top-10 nearest tokens are toxic.
We adopt LLM-as-a-judge \cite{zheng2023judgingllmasajudgemtbenchchatbot} using GPT-4o \cite{openai2024gpt4ocard} to evaluate whether a token is considered toxic (e.g. curse words, slurs, sexual content). 
See Appendix Table~\ref{tab:logit_lens_toxic} for the tokens projected by these toxic value vectors.

We then counterfactually isolate their effect on toxicity scores using activation patching (Section~\ref{subsec:activation_patching_setup}).
Namely, for each pre-trained model, we set the activations of toxic value vectors to that of its post-DPO counterpart.

% As an additional baseline, we also forbid each LLM generating the tokens projected by those toxic neurons.
% Table~\ref{tab:forbidden_toxic_tokens} lists the tokens excluded which are projected to by toxic neurons across models.

Table~\ref{tab:num_toxic_neurons} reports the number of toxic neurons per model and the percentage reduction in toxicity scores through patching.
Toxic neurons comprise fewer than 0.05\% of all MLP neurons, and account for as little as 2.5\% to 24\% of the reduction in toxicity scores depending on the model. 
As patching captures interactions between toxic and non-toxic neurons, 
these results suggest that toxic neurons only account for a small portion of DPO’s effect, rendering \citet{lee2024mechanisticunderstandingalignmentalgorithms}'s claim that DPO primarily dampens toxic neurons as incomplete.

\begin{table}[ht]
\centering
\small
\setlength{\tabcolsep}{4pt} 
\caption{\textit{The number of toxic neurons per model and percentage decrease in toxicity scores after patching them.} 
% We report both the toxic neurons with toxic tokens and the top 256 neurons most aligned with the probe. 
The first row reports the number of toxic neurons with toxic tokens. 
The second row reports the top 256 toxic-aligned neurons. 
The percentage decrease is the proportion of toxicity score reduction from patching toxic neurons, relative to the total reduction of DPO 
(see Table~\ref{tab:all_patching_results_main} for the full scores).}
\begin{tabular}{p{1.7cm} p{1.7cm} p{1.7cm} p{1.7cm}}
\toprule
\makecell{\textbf{GPT-2}\\\textbf{355M}} & \makecell{\textbf{Llama}\\\textbf{3.1-8B}} & \makecell{\textbf{Gemma}\\\textbf{2-2B}} & \makecell{\textbf{Mistral}\\\textbf{7B}} \\
\midrule
59 (19.7\%$\downarrow$) & 7 (1.96\%$\downarrow$) & 3 (0.41\%$\downarrow$) & 14 (0.63\%$\downarrow$) \\
256 (23.9\%$\downarrow$) & 256 (3.14\%$\downarrow$) & 256 (2.47\%$\downarrow$) & 256 (16.5\%$\downarrow$) \\
\bottomrule
\end{tabular}
\label{tab:num_toxic_neurons}
\end{table}

\begin{figure*}[t!]
    \centering \includegraphics[width=0.96\textwidth]{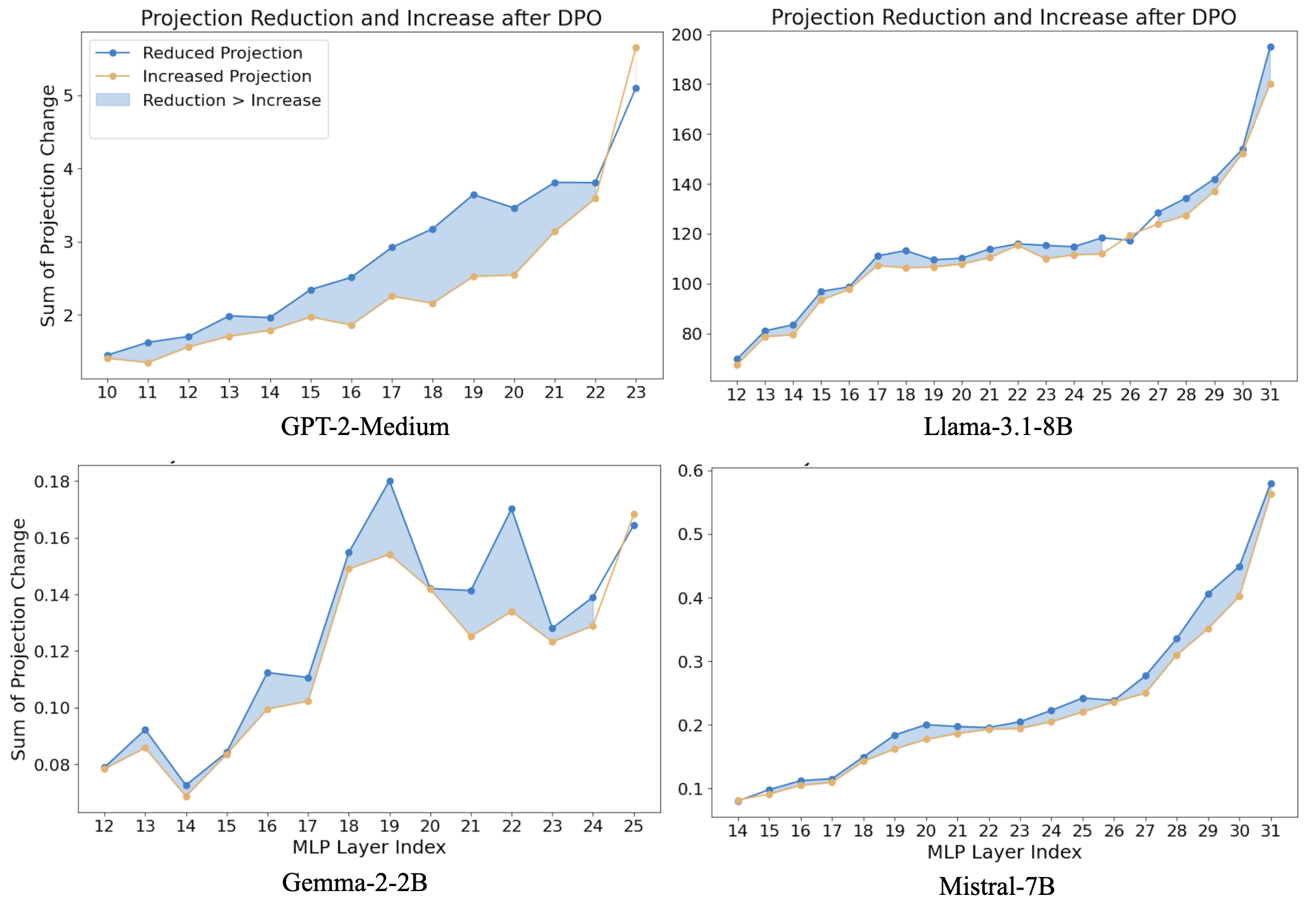}
   \caption{\textit{  
   DPO balances opposing toxicity writing across MLP layers.} 
   % leading to a net reduction
   % through distributed activation shifts 
   Blue dots show total projection reduction per layer, 
   orange dots show the total increase, 
   both after DPO. 
   The shaded blue areas illustrate how these opposing effects cancel out and lead to a net toxicity reduction. 
   Projection changes grow with layers when measured against last-layer probe.
   Net changes in first $\approx10$ layers are negligible and omitted; 
   see Appendix Table~\ref{fig:four_model_net_reduction_full} for the full graph. }
   %  with all layers
   % which is most pronounced in later MLP layers. 
   \label{fig:four_model_net_reduction}
\end{figure*}

\section{A Deeper Look at DPO Weight Shifts}
\label{sec:deeper_look_at_weight_shifts}
Here, we show that the weight shifts from DPO are more nuanced than simply bypassing toxic neurons.

\subsection{DPO Balances Opposing Effects}
\label{subsec:balance_opposing_effects}

Across all models, DPO makes minimal adjustments to the MLP weights. 
All MLP value vectors have a cosine similarity of 0.99 before and after DPO, likely due to the KL divergence regularisation~\cite{rafailov2024directpreferenceoptimizationlanguage}. 
However, these small weight changes ($\mathbf{v}_i^{\text{pre}} \approx \mathbf{v}_i^{\text{dpo}}$) accumulate and induce distributed activation shifts ($m_i^{\text{pre}}-m_i^{\text{dpo}}$) across \textit{all} MLP neurons.
The majority of neurons undergo average shifts ranging from 0.66\% (Llama-3.1-8B) to 16.71\% (Mistral-7B), with substantial variation in the tails (see Appendix Figure~\ref{fig:four_model_act_dist}).
% small activation shifts, with 

% \begin{table}[ht]
% \centering
% \small
% \setlength{\tabcolsep}{4pt} 
% \caption{\textit{Averaged activation shifts during DPO.} The averaged shifts varies across models. }
% \begin{tabular}{cccc}
% \toprule
% \makecell{\textbf{GPT-2-355M}} & \makecell{\textbf{Llama-3.1-8B}} & \makecell{\textbf{Gemma-2-2B}} & \makecell{\textbf{Mistral-7B}} \\
% \midrule
% 3.11\% & 0.66\% & 11.80\% & 16.71\% \\
% \bottomrule
% \end{tabular}
% \label{tab:avg_act_shifts}
% \end{table}

% these activation shifts are largely random. 
% Pearson correlation analysis \cite{schober2018correlation} reveals no correlation between these activation shifts and the ``toxicity level'' of a neuron, measured by its cosine similarity with the toxic probe, and only a weak positive correlation with pre-trained activations (see Appendix Table~\ref{tab:act_correlation_analysis}). 

These distributed activation shifts lead approximately half of all neurons (52\%$\sim$58\% across models) reducing their projection onto the toxic direction ($\Delta_{\text{Toxic}, i} > 0$) and the other half increasing it ($\Delta_{\text{Toxic}, i} < 0$) 
(see Appendix Table~\ref{tab:percentages_increase_decrease}). 
Figure \ref{fig:four_model_net_reduction} illustrates how the  opposing neuron effects accumulate and balance out at each MLP layer, resulting in a net toxicity reduction. 
This suggests that DPO does not simply suppress toxic signals, but rather delicately redistributes them, balancing a trade-off across all MLP neurons.

\begin{figure*}[t!]
    \centering
    \includegraphics[width=0.96\textwidth]{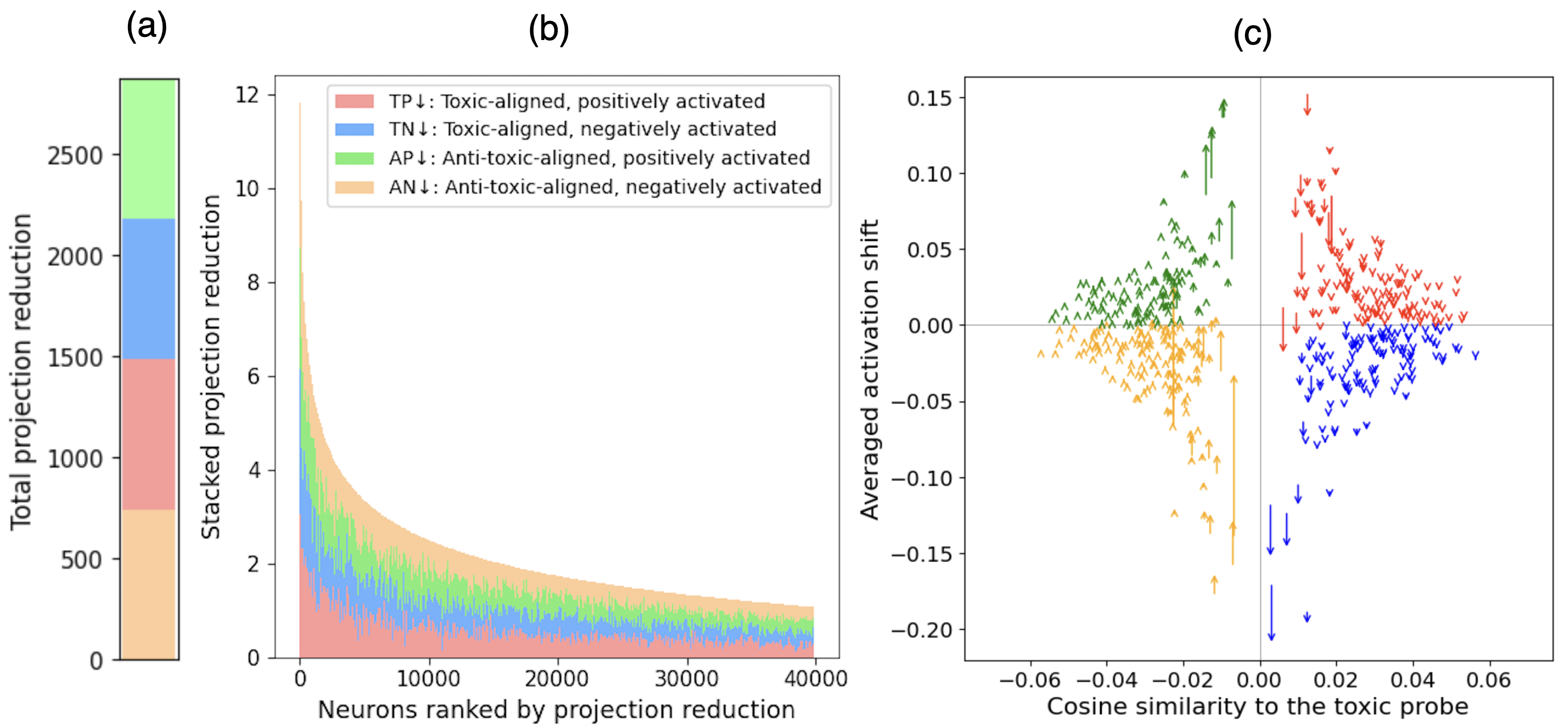}
    \vspace{-2mm}
 \caption{
\textit{Four neuron groups collectively reduce toxicity during DPO, shown for Llama-3.1-8B.} 
The same four groups emerge consistently across models, with panels (a) and (b) showing slightly different patterns for the other three models (see Appendix Figure~\ref{fig:neuron_group_contributions_all_models}). 
(a) Proportion of toxicity reduction per group, showing balanced contributions;
(b) Cumulative toxicity reduction for top 40,000 neurons (ranked by reduction in projection), where groups show similar reduction rates;
(c) Per-group activation shifts during DPO for the top 2,000–2,500 neurons, where each group shifts according to their orientation relative to the toxic representation.}
\label{fig:neuron_group_contributions}
\end{figure*}

\subsection{Four Neuron Groups Reduce Toxicity}
% Collectively
\label{subsec:neuron_groups}
Based on these results, 
we study value vectors that \textit{reduce} toxic projections ($\Delta_{\text{Toxic}, i} > 0$), as they likely contribute to toxicity reduction during DPO. 
We categorise them into four mutually exclusive groups, and study their collective effect.

Table~\ref{tab:neuron_groups_def} defines the four neuron groups, categorised by their alignment with the toxicity probe (\textbf{T}oxic-aligned vs. \textbf{A}nti-toxic-aligned) and their pre-DPO activations (\textbf{P}ositive vs. \textbf{N}egative).
% , \TP{}, \TN{}, \AP{}, \AN{}
Namely, 
\TP{}, \TN{} have positive alignment with toxicity, while \AP{}, \AN{} have negative alignment.
All groups reduce toxicity projection during DPO ($\downarrow$).
Table~\ref{tab:proportion_four_groups} shows the proportions of neurons in each group across models.
Note that \citet{lee2024mechanisticunderstandingalignmentalgorithms} only considers the neurons in \TP{}.

Figure~\ref{fig:neuron_group_contributions}c visualises how these four groups reduce toxicity writing via activation shifts for Llama-3.1-8B, with similar patterns observed in all models (see Appendix Figure~\ref{fig:neuron_group_contributions_all_models}).
The activations of each group are shifted in accordance to their orientation with respect to the toxic probe.
Namely, toxic-aligned weights (\TP{}, \TN{}) drop in activations, while anti-toxic aligned weights (\AN{}, \AP{}) increase in activations (promotion of ``anti-toxicity'').

\begin{table}[ht]
    \centering
    \small
    \caption{\textit{Definitions of four neuron groups reducing toxicity projections} ($\Delta_\text{Toxic, i} > 0$).
    \textit{Alignment with probe} (T vs. A) indicates whether the neuron’s value vector $\mathbf{v}$ aligns positively or negatively with the toxic probe \( W_{\text{Toxic}} \) ($\mathbf{v} \cdot W_{\text{Toxic}} > 0$ or $\mathbf{v} \cdot W_{\text{Toxic}} < 0$).
    }
    \begin{subtable}[t]{1\linewidth}
        \centering
        \begin{tabular}{cccc}
            \toprule
            \textbf{Group} &
            \makecell[c]{\textbf{Alignment} \\ \textbf{with probe}} &
            \makecell[c]{\textbf{Pre-DPO} \\ \textbf{activation}} &
            \makecell[c]{\textbf{Projection} \\ \textbf{change}} \\
            \midrule
            \textcolor{red}{TP $\downarrow$}   & \textbf{T}oxic-aligned      & \textbf{P}ositive & Reduced (↓)\\
            \textcolor{blue}{TN $\downarrow$}  & \textbf{T}oxic-aligned      & \textbf{N}egative & Reduced (↓)\\
            \textcolor{teal}{AP $\downarrow$} & \textbf{A}nti-toxic-aligned & \textbf{P}ositive & Reduced (↓)\\
            \textcolor{orange}{AN $\downarrow$}& \textbf{A}nti-toxic-aligned & \textbf{N}egative & Reduced (↓)\\
            \bottomrule
        \end{tabular}
    \end{subtable}
    \hfill
    \\[0.8em]
    %\begin{subtable}[t]{1\linewidth}
    %    \centering
    %    \caption*{(b) \textit{Two criteria for defining neuron groups. }}
    %    \begin{tabular}{@{}p{1.5cm} p{6cm}@{}}
    %        \toprule
    %        \textbf{Criterion} & \textbf{Description} \\
    %        \midrule
    %        Alignment with probe 
    %        % (\textbf{T} or \textbf{A})
    %        & Whether the neuron’s value vector \( v_i \) aligns positively or negatively with the toxic probe \( W_{\text{Toxic}} \) ($v_i \cdot W_{\text{Toxic}} > 0$ or $v_i \cdot W_{\text{Toxic}} < 0$). \\[0.8em]
    %        Pre-DPO activation %(\textbf{P} or \textbf{N}) 
    %        & Whether the neuron is positively or negatively activated before DPO (\( m_i^{\text{pre}} > 0 \) or \( m_i^{\text{pre}} < 0\)). \\
    %        % Projection change ($\uparrow$ or $\downarrow$) & Whether the neuron's toxicity projection increases (\textbf{↑}) or reduces (\textbf{↓}) after DPO ($\Delta_\text{Toxic, i} > 0$ or $\Delta_\text{Toxic, i} < 0$). \\
    %        \bottomrule
    %    \end{tabular}
    %\end{subtable}
    \label{tab:neuron_groups_def}
\end{table}

\vspace{-3mm}
\begin{table}[ht]
    \centering
    \small
\caption{\textit{Proportions of four-neuron-group among all neurons reducing toxicity projection ($\downarrow$).} 
    Proportions are more balanced across larger LLMs. 
    The \textit{Sum} column shows the total number of neurons per model.}
    \begin{tabular}{@{}p{1.8cm} p{0.7cm} p{0.7cm} p{0.7cm} p{0.7cm} p{1.0cm}@{}}
        \toprule
        \textbf{Model} & 
        \textcolor{red}{$\rm TP\downarrow$} & 
        \textcolor{blue}{$\rm TN\downarrow$} & 
        \textcolor{teal}{$\rm AP\downarrow$} & 
        \textcolor{orange}{$\rm AN\downarrow$} & 
        \textbf{Sum} \\
        \midrule
        GPT-2-355M     & 6.9\%  & 39.1\% & 3.2\%  & 50.9\% & 57,501 \\
        Llama-3.1-8B  & 25.4\% & 24.4\% & 24.6\% & 25.5\% & 239,460 \\
        Gemma-2-2B    & 28.8\% & 21.3\% & 21.3\% & 28.6\% & 123,898 \\
        Mistral-7B    & 29.7\% & 20.3\% & 20.2\% & 29.8\% & 238,236 \\
        \bottomrule
    \end{tabular}
\label{tab:proportion_four_groups}
\end{table}

\paragraphx{Anti-toxic value vectors.}
What do ``anti-toxic'' value vectors encode?
Geometrically, some anti-toxic value vectors essentially lie at the antipode of toxic semantic clusters. 
% in the representation space.
Namely, we take value vectors with highest  cosine similarity scores to $-1 \times W_\text{Toxic}$ (i.e. anti-toxic).
We then multiply these value vectors by $-1$, unembed them, and inspect their nearest tokens.
Table~\ref{tab:logit_lens_anti_toxic} shows examples of toxic tokens they project to (see Appendix Table~\ref{tab:logit_lens_anti_toxic_appendix} for more examples).
This shows how DPO promotes anti-toxicity by increasing the activation of anti-toxic \AN{}, \AP{} neurons.

\begin{table}[t!]
\centering
\renewcommand{\arraystretch}{1.3}
\arrayrulecolor{black}
\small
\vspace{-0.5em}
\caption{\textit{Examples of anti-toxic value vectors (with reversed signs) that project to toxic tokens via Logit Lens.} 
\textcolor{red}{Warning: these examples are highly offensive.}}
\label{tab:logit_lens_anti_toxic}

\hspace{30mm}

\begin{tabular}{@{}p{0.8cm} p{1.6cm} p{4.5cm}@{}}
%\begin{tabular}{lll}
\toprule
\textbf{Model} & \textbf{Vector} & \textbf{Top tokens} \\
\midrule
GPT2 & $-1\!\times\!\mathbf{v_{11}^{1307}}$ & d*mn, darn, kidding, freaking, piss \\ 
Llama3 & $-1\!\times\!\mathbf{v_{25}^{14671}}$ & f*ck, f*cked, f*cking, sh*t, F*CK \\ 
%Gemma2 & $-1\!\times\!\mathbf{v_{14}^{7822}}$ & f*cking, godd*mn, f*ck, sh*t, d*mn\\ 
Gemma2 & $-1\!\times\!\mathbf{v_{14}^{7822}}$ & f*cking, godd*mn, f*ck, sh*t \\ 
Mistral & $-1\!\times\!\mathbf{v_{14}^{14693}}$ & sh*t, f*ck, Block, piss, f*cking \\

\bottomrule
\end{tabular}
\end{table}

\paragraphx{Why negative activations?}
% Origin of negatively activated neurons
Negatively activated neurons (including \TN{}, \AN{}) take a large portion of MLP neurons, around 50\% in three larger models and 87\% in GPT-2 Medium 
(see Appendix Table~\ref{tab:percentages_neurons_neg_act}). 
This results from the modern choices of activation functions: GeLU (GPT-2), GeLU-Tanh (Gemma), and SiLU (Llama, Mistral), 
which allow neurons to retain small negative activations for negative inputs 
\cite{hendrycks2023gaussianerrorlinearunits}.  
This enables plenty of neurons to maintain gradient flow and contribute marginally to the toxicity representation through their activation shifts.
%  minimal
% rather than relying on a small set of strongly active neurons as with ReLU. 

%Figure~\ref{fig:neuron_groups_stacked_bar} shows that \textcolor{red}{$\rm TP_{-}$} and \textcolor{orange}{$\rm AN_{-}$} are the primary contributors, accounting for 69.1\% of the total reduction by removing existing toxicity, and the other two groups contribute 30.9\% by promoting anti-toxicity. 

\paragraphx{Four groups reduce toxicity at different rates.}
When ranking neurons by their reduction of toxicity projection, the four groups show different reduction rates.
In Llama-3.1-8B, all groups contribute evenly, maintaining balanced shares of top-ranked neurons 
(Figure~\ref{fig:neuron_group_contributions}b).
% This results in balanced sum of toxicity reduction across groups (Figure~\ref{fig:Llama3_neuron_groups_stacked_bar}).
% This results in balanced sums of reduction across groups (Figure \ref{fig:Llama3_neuron_groups_stacked_bar}).
In contrast, in the other three models, 
\textcolor{red}{$\rm TP\downarrow$} dominating the top-ranked neurons, while \textcolor{orange}{$\rm AN\downarrow$} gradually gains influence in later ranks—a trend most evident in GPT-2-Medium 
(see Appendix Figure~\ref{fig:neuron_group_contributions_all_models}).
As a result, \textcolor{red}{$\rm TP\downarrow$} and \textcolor{orange}{$\rm AN\downarrow$} dominate their overall toxicity reduction. 
% in these models.

% We further demonstrate how the four groups accumulate toxicity reduction across MLP layers in all models.
\vspace{-0.5mm}
\paragraphx{Reduction peaks in later layers.}
We also observe an overall increasing trend in toxicity reduction across MLP layers (see Appendix Figure~\ref{fig:per_layer_neuron_groups}). 
This shows that the four groups collectively steer each layer away from toxicity, with later layers giving the strongest suppression of toxic outputs.
This upward trend may be partly due to the probes being extracted from the final layer.
% . As before, \TP{} and \AN{} dominate toxicity reduction across layers and models.

\paragraphx{Activation patching confirms the collective effects of four groups.}
Finally, we confirm the collective effect of the four groups using activation patching.
This post-hoc analysis assumes access to each group's activations after DPO and evaluates their effects counterfactually by patching each neuron group, one at a time, in the pre-trained model to match their post-DPO activations.

Table~\ref{tab:all_patching_results_main} shows that sequentially patching each group further reduces toxicity scores across all models, confirming each neuron group’s contribution to DPO's effects. 
% This confirms the contributions of both anti-toxic and negatively activated groups to DPO’s effects. 
Furthermore, patching all four groups either surpasses or closely matches DPO’s toxicity reduction and consistently outperforms probe-based steering.
It has minimal impact on perplexity and only slightly reduces F1 scores across models.
This patching outperforms DPO likely because it excludes neurons that increase toxicity projection after DPO (Section~\ref{subsec:balance_opposing_effects}). 
As a sanity check, patching all neurons that increase toxicity projection (↑) during DPO leads to higher toxicity scores across models, consistent with the projection changes 
(see Appendix Table~\ref{tab:all_patching_results}).

\section{Activating Editing to Replicate DPO}
\label{subsec:replicating_dpo_with_activation_editing}
Based on our insights, 
we demonstrate two simple methods to replicate DPO's effects by directing editing activations. 
These methods only rely on a toxicity representation (e.g. a probe) and do not require any weight updates nor a pairwise preference dataset, which is not always readily available.
Unlike the previous activation patching analyses, here we do not assume access to post-DPO activations.
% weights or 

\subsection{Probe-based Activation Editing}
Previously, we focused on neuron groups with reduced toxicity projections (i.e., $\Delta_\text{Toxic, i} > 0$) (Section~\ref{subsec:neuron_groups}).
However, knowing whether a neuron increases or decreases in toxicity projection requires access to post-DPO activations (see Equation~\ref{eq:toxicity_reduction}). 
To remove this dependency, we re-categorise the neuron groups based solely on their alignment with the toxicity probe and their pre-DPO activations, 
and do not consider their projection changes (hence notated as \TPbroad{} as opposed to \TP{}).

Given our new neuron groups (\TPbroad{}, \TNbroad{}, \APbroad{}, \ANbroad{}), we leverage two key insights learned from DPO: 
activation shifts are distributed across all neurons (Section~\ref{subsec:balance_opposing_effects}), and the direction of activation shifts for toxicity reduction depends on the orientation of the value vector (Section~\ref{subsec:neuron_groups}, Figure~\ref{fig:neuron_group_contributions}c).

Follow these insights, we sample a fraction  $\beta$ (\%) of neurons from each group and minimally adjust their activations. 
For toxicity-aligned groups (\TPbroad{}, \TNbroad{}), 
we slightly decrease their activations by a factor of $\alpha$ (\%), 
while for anti-toxicity-aligned groups (\APbroad{}, \ANbroad{}) we slightly increase them. 
As \TNbroad{} and \ANbroad{} have negative activations, we flip the sign of $\alpha$ accordingly:
% to adjust their proportions
\begin{equation*}
\label{eq:activation_editing}
\begin{aligned}
m_{\textcolor{red}{\mathrm{TP}_\beta}}^{\text{edit}} &\!=\! (1 \!-\! \alpha) m_{\textcolor{red}{\mathrm{TP}_\beta^\text{}}}^\text{pre} ;\quad m_{\textcolor{blue}{\mathrm{TN}_\beta}}^{\text{edit}} = (1 \!+\! \alpha) m_{\textcolor{blue}{\mathrm{TN}_\beta}}^\text{pre} \\
m_{\textcolor{teal}{\mathrm{AP}_\beta}}^{\text{edit}} &\!=\! (1 \!+\! \alpha) m_{\textcolor{teal}{\mathrm{AP}_\beta}}^\text{pre};\quad
m_{\textcolor{orange}{\mathrm{AN}_\beta}}^{\text{edit}} = (1 \!-\! \alpha) m_{\textcolor{orange}{\mathrm{AN}_\beta}}^\text{pre}
%m_{\textcolor{red}{\mathrm{TP}_\beta}}^{\text{edit}} &= (1 - \alpha) \cdot m_{\textcolor{red}{\mathrm{TP}_\beta^\text{}}}^\text{pre} \\
%m_{\textcolor{blue}{\mathrm{TN}_\beta}}^{\text{edit}} &= (1 + \alpha) \cdot m_{\textcolor{blue}{\mathrm{TN}_\beta}}^\text{pre} \\
%m_{\textcolor{teal}{\mathrm{AP}_\beta}}^{\text{edit}} &= (1 + \alpha) \cdot m_{\textcolor{teal}{\mathrm{AP}_\beta}}^\text{pre}
%\\
%m_{\textcolor{orange}{\mathrm{AN}_\beta}}^{\text{edit}} &= (1 - \alpha) \cdot m_{\textcolor{orange}{\mathrm{AN}_\beta}}^\text{pre}
\end{aligned}
\end{equation*}
where \( \mathrm{TP}_\beta \), \( \mathrm{AN}_\beta \), \( \mathrm{TN}_\beta \), and \( \mathrm{AP}_\beta \) denote the \( \beta \)\-fraction of neurons in each group, and \( m^{\text{pre}} \) are their pre-trained activations.
Again, here we do not rely on any post-DPO information (i.e., \(m^\text{DPO}\)).

Table~\ref{tab:all_patching_results_main} shows the results for selected hyperparameters $\alpha$ and $\beta$.
These hyperparameters reflect our insights: 
most neurons (high $\beta$ value) undergo  small shifts (small $\alpha$ value).
We find that selecting the top-$\beta$ fraction of neurons ranked by cosine similarity with the toxicity probe is most effective in reducing toxicity scores. 
In particular, selecting $\beta = 55\%$ provides the best trade-off between toxicity reduction and F1 preservation, 
consistent of our earlier finding that DPO reduces toxicity writing in roughly half of all neurons (Section~\ref{subsec:balance_opposing_effects}).
%Selecting $\alpha = 0.01$ corresponds to the lower end of the average activation shifts observed across models (Section~\ref{subsec:balance_opposing_effects}).  % (see Appendix Figure~\ref{fig:four_model_act_dist}). 
This approach outperforms both DPO and probe-based steering in toxicity reduction while preserving perplexity across pre-trained models, 
with only a slight F1 score  decrease. 
%It also achieves better perplexity than both DPO and probe-based steering.
Further increasing $\beta$ (e.g., to 0.8) leads to greater toxicity reduction at the cost of F1 drops.
Alternative sampling strategies for selecting the top-$\beta$ neurons (e.g., by ascending absolute activation values) yield similar toxicity reduction across models (see Appendix Table~\ref{tab:all_patching_results}). 
% In contrast, selecting $\beta$ by descending absolute activation values tends to severely degrade perplexity, particularly in GPT-2 Medium and Llama-3.1-8B.

\subsection{Probe-free Activation Editing}
While the previous activation editing method does not require pairwise preference data, it still relies on a latent toxicity representation, for which we use our probe. 
While a probe does not require pairwise preference data, 
it still requires labelled classification data  (Section~\ref{subsec:per_neuron_toxicity_contributions}).

Here, we demonstrate that activation editing can be performed even without a probe by leveraging an alternative toxicity representation. 
Namely, prior works have observed a close relationship between concept representations in the model's hidden layers and the token embedding space~\cite{lee2025sharedgloballocalgeometry}.
Similarly, we find that toxic tokens are nearest neighbors to our probes in the token embedding space (Table~\ref{tab:logit_lens}).
Motivated by this, we replace the probe with a contrastive vector derived directly from token embeddings.

To construct this vector, we simply select sets of toxic and non-toxic token embeddings for each model and compute the difference between their mean embeddings (Table~\ref{tab:toxic_nontoxic_tokens}).
This bypasses the need to train a probe model.
We then apply the same activation editing method as described above.

\begin{table}[ht]
\centering
\small
\renewcommand{\arraystretch}{1.3}
\setlength{\tabcolsep}{4.7pt} 
\caption{\textit{Toxic and non-toxic tokens for computing the contrastive vector.} The contrastive vector is obtained by subtracting the mean embedding of non-toxic tokens from that of toxic tokens.}
\label{tab:toxic_nontoxic_tokens}
\begin{tabular}{llllll}
\toprule
\textbf{Toxic} & \texttt{fu*k} & \texttt{sh*t} & \texttt{cr*p} & \texttt{da*n} & \texttt{a**hole}\\
\textbf{Non-toxic} & \texttt{hello} & \texttt{thanks} & \texttt{friend} & \texttt{peace} & \texttt{welcome} \\
\bottomrule
\end{tabular}
\end{table}

The last rows of Table~\ref{tab:all_patching_results_main} show that this probe-free approach achieves results comparable to the probe-based method. 
Together, these results validate our understanding of DPO and offer a proof-of-concept alternative when weight updates are prohibitively costly or training data is not readily available.

%\begin{table}[ht]
%    \centering
%    \small
%    \caption{\textit{Activation editing: description of parameters.}}
%    \begin{tabular}{@{}p{0.5cm} p{6.5cm}@{}}
%        \toprule
%        \textbf{} & \textbf{Description} \\
%        \midrule
%        $\alpha$ & The percentage of activation shift on pre-trained activation $m_{i}^{pre}$.\\[0.8em]
%        $\beta$ & The proportion of neurons to intervene in each broader neuron group \textcolor{red}{$\rm TP$}, \textcolor{orange}{$\rm AN$}, \textcolor{blue}{$\rm TN$}, \textcolor{green}{$\rm AP$}. \\
%        \bottomrule
%    \end{tabular}
%    \label{tab:editing_alpha_beta}
%\end{table}

% Perplexity and F1 depends on the number of neurons intervening on (e.g. $\rm TP\downarrow$ has much less perplexity break than TP) and the amount of activation shifts (1.5 or 1.25).

\section{Discussion and Conclusion} 
Our work provides a mechanistic understanding of how DPO reduces toxicity across four LLMs.
Using activation patching, we showed that prior explanations are incomplete \cite{lee2024mechanisticunderstandingalignmentalgorithms}: a small set of toxic neurons associated with toxic tokens cannot fully account for DPO’s effects. 
This explanation also relies on a monosemantic view of neurons, 
an assumption disputed by prior work \cite{elhage2022toymodelssuperposition}.
Instead, DPO induces distributed activation shifts across all MLP neurons, leading to a net reduction in toxicity.

To characterise these distributed effects, 
we identified four neuron groups that play distinct roles in toxicity reduction and show that their combined effect replicates that of DPO. 
Building on these insights, we developed an activation editing method that mimics DPO by applying distributed activation shifts along a learned toxicity representation.
We explored two options for this representation: a probe model and a contrastive vector derived from token embeddings.
This method outperforms DPO in reducing toxicity while preserving perplexity, all without any weight updates. 
%Its success further validates our mechanistic understanding of DPO. 
 
% to safety fine-tuning. 
% mechanistic 

% if so, the wrapper is very spreaded across many neurons (if not all neurons), contridicting to Jain 2023 findings that wrappers are very localised - this is discussed by Lee due to the KL term in RLHF's loss function to discourage parameter changes

% comment on DPO does more than learning an offset

%\subsection{The Shallowness of Safety} 
DPO's tendency to spread activation shifts thinly across the network suggests that pre-trained harmful capabilities are merely thinly masked. 
% not removed, but 
As a result, small disruptions anywhere in the model, not just in toxic neurons, 
can potentially breach the safety barrier and reactivate harm. 
This extends prior findings on the shallowness of safety fine-tuning from the activation perspective \cite{jain2024makesbreakssafetyfinetuning, qi2024safetyalignmentjusttokens}.
These distributed shifts likely arise as a by-product of regularisation to preserve pre-training performance, 
hinting at a deeper trade-off: the shallow safety may be an inherent cost of maintaining language quality. 
This diluted effect is further compounded by smooth activation functions (Section~\ref{subsec:neuron_groups}), which allow many weakly active neurons to marginally contribute to  toxicity writing.
This leaves much of the model’s toxic capacity  untapped. 
In fact, many MLP neurons increase their toxicity projection during DPO (Section~\ref{subsec:balance_opposing_effects}). 
In contrast, our activation editing method offers a more targeted alternative  by explicitly steering activations to reduce toxicity.  
This may explain why it achieves greater toxicity reduction than DPO, 
despite applying smaller average activation changes. 
Taken together, our findings point to the value of exploring more interpretable safety interventions as a path beyond shallow tuning. 

% \paragraphx{Why patching all four does not always outperform DPO} 
% Since we exclude neurons adding projection, it supposes to reduce more than DPO, which it did for gpt and mistral, but:
% down neurons may evoke up neurons as projection values do not account for interactions

In summary, our work provides a more complete understanding of how DPO reduces toxicity and introduces an efficient, training-free alternative.

\section{Limitations}
\label{sec:limitations}
\paragraphx{Projection to a toxic subspace.} 
In this work, we use a linear probe to capture an aggregated toxicity representation, following common practice in the literature \cite{ferrando2024primerinnerworkingstransformerbased, ravfogel22a}. 
However, it may be possible that toxicity manifest along multiple directions, each capturing different aspects such as hate speech or abusive language, and thus better represented as a subspace \cite{uppaal2024detoxtoxicsubspaceprojection}. 
We conduct an initial analysis on GPT-2-Medium and find that using a subspace complicates our identification of neuron groups. 
We construct a toxic subspace via Singular Value Decomposition (SVD) on the top 128 toxic-aligned value vectors, where each of the top three singular vectors projects to different toxic tokens 
(see Appendix~\ref{appendix:why_not_toxic_subspace}).
We find that most value vectors show inconsistent alignment across the three directions and mixed projection changes to the toxicity probe after DPO. 
A single value vector can be ``toxic-aligned'' in one SVD direction and ``anti-toxic-aligned'' in another, reducing toxicity along one axis while increasing it in another.  
These inconsistencies make it difficult to assign neurons to coherent groups as in our approach.
We therefore leave a more robust analysis of toxic subspace projections to future work. 

% This also means we have four separate neuron groups for each SVD direction, where the neuron groups actually contradict each other. 

% Without reliable groupings, it is difficult to link toxicity scores to the identified neuron groups via patching and verify their effects. 

% the low cossim between probe and toxic embedding suggest toxic representations can indeed embed in very different directions; also for SVD vectors their cossim should be zero/near zero for truncated SVD

\paragraphx{Assumptions for projection.} 
We use projections to estimate each neuron's contribution to toxicity (Equation~\ref{eq:toxicity_reduction}), assuming that neurons contribute proportionally along their activated directions. However, toxicity representations may be distributed across more complex linear combinations of neurons. Alternative tools, such as sparse autoencoders (SAEs) \cite{bricken2023towards, cunningham2023sparseautoencodershighlyinterpretable}, which learn linear feature compositions through autoencoder reconstruction, may offer a complementary perspective for tracing toxicity feature changes back to specific neurons.

\paragraphx{Generalise the four neuron groups across tasks and models.} 
% attention blocks 
% % Extend analysis to attention layer outputs, also measure Q,K,V,O vector shifts corrsponding ti the probe and contribution?
DPO is inherently a binary algorithm, designed to train on pairwise preference data. The four neuron groups we identify naturally reflect this binary structure, where we find that their activations shift  along the representation of a binary concept. We therefore expect similar neuron group structures to emerge in other binary safety-related tasks trained with DPO beyond toxicity (e.g., biased vs. unbiased content, factual vs. misinformation), a direction we leave for future work.

These neuron groups may also persist in general instruction-tuned models (e.g., those trained with supervised fine-tuning or RLHF) on binary tasks, likely also  operating through distributed activation shifts due to regularisation. 
We leave this as another direction for exploration.

\paragraphx{Generalise the activation editing method to more tasks.}  
Our activation editing method requires only a linear concept representation, which can be derived from a probe or token embeddings—both relatively cheap to obtain. 
%without needing weight updates, pairwise preference data, or even classification data.
Future work could extend our method to other safety-related tasks (e.g., bias or misinformation) where such representations can be derived from classification data, or to general tasks where the target behaviour can be captured by representative tokens (e.g., sentiment polarity, political stance).

\bibliography{custom}
\clearpage
\appendix

\addcontentsline{toc}{section}{Appendix} % Add the appendix text to the document TOC
\part[]{}
\parttoc % Insert the appendix TOC

\section{Gated Linear Units}
\label{appendix:glu}
In this section, 
we introduce Gated Linear Units (GLUs), which replace standard MLPs (Section~\ref{sec:related_work}) in recent models such as Llama, Gemma, Mistral \cite{shazeer2020gluvariantsimprovetransformer}.

GLUs introduce a gating mechanism that selectively controls information flow by computing the element-wise product of two linear projections, one of which is passed through a non-linearity $\sigma$:
\[
\text{GLU}^{\ell}(\mathbf{x}^{\ell}) = \Big(\sigma(W_1^{\ell} \mathbf{x}^{\ell}) \odot W_2^{\ell} \mathbf{x}^{\ell}\Big) W_V^{\ell},
\]
where $W_1^{\ell}, W_2^{\ell}, W_V^{\ell} \in \mathbb{R}^{d_{mlp} \times d}$.
The term \( \sigma(W_1^\ell \mathbf{x}^\ell) \) acts as the \textit{gates}, blocking \( W_2^\ell \mathbf{x}^\ell \) from propagating when the non-linearity (\(\sigma\)) is inactive.

We can still express GLUs as (see Equation~\ref{eq:mlp_exp}):
\[
\text{MLP}^{\ell}(\mathbf{x}^{\ell}) = \sum_{i=1}^{d_{\text{mlp}}} m_i^{\ell} \mathbf{v}_i^{\ell},
\]
where 
\[
    m_i^{\ell} = \sigma(\mathbf{k}_i^{\ell} \cdot \mathbf{x}^{\ell}) \cdot (\mathbf{w}_i^{\ell} \cdot \mathbf{x}^{\ell}),
\]
$\mathbf{k}_i^{\ell} \in \mathbb{R}^{d}$ and $\mathbf{w}_i^{\ell} \in \mathbb{R}^{d}$ are the $i$-th rows of $W_1^{\ell}$ and $W_2^{\ell}$, respectively. 
For each MLP neuron $i$, $\mathbf{v}_i^\ell$ (rows of $W_V^{\ell}$) is its \textit{value vector} \cite{geva2021transformerfeedforwardlayerskeyvalue}, 
and the scalar 
\( m_i^{\ell} \in \mathbb{R} \) is an \textit{activation score} that controls the scaling of the value vector \( \mathbf{v}_i^\ell \). 
% This means an MLP layer writes to the residual stream $d_{\text{mlp}}$ times, once per neuron, via the activation-weighted value vectors $m_i^{\ell} \mathbf{v}_i^{\ell}$. 
% $\mathbf{k}_i^{\ell}$ (the $i$-th row of $W_K^{\ell}$) as the \textbf{key vector}, and 

This shows that, despite despite architectural differences in GLUs, our formulation in Equation~\ref{eq:mlp_exp} still holds, as it consists of value vectors scaled by a non-linear activation.

% \newpage
\section{MLP layer specification}
\label{appendix:mlp_spec}
In this section, we provide the MLP layer specifications for each model (Section~\ref{subsec:data_and_mdoels}).

Table~\ref{tab:model_mlp_specs} reports, for each model, the number of MLP layers, MLP hidden dimensions, activation function, and whether a gating mechanism is used.

\begin{table}[ht]
    \centering
    \caption{\textit{MLP specifications for each model.} 
    $l$ is the number of MLP Layers, 
    $d$ is the residual stream dimension, $d_{\text{mlp}}$ is the dimension of MLP hidden layer, $\sigma$ is the activation function, \textit{Gated?} indicates whether the model uses gated MLPs.}
    \renewcommand{\arraystretch}{1.2} % Increase row spacing
    \small
    \begin{tabular}{@{}l c c c c c@{}}
        \toprule
        \textbf{Model} & \textbf{$l$} & \textbf{$d$} & \textbf{$d_{\text{mlp}}$} & \textbf{$\sigma$} & \textit{Gated?} \\ 
        \midrule
        GPT-2-355M  & 24  & 1024  & 4096  & GeLU & ✗ \\ 
        Llama-3.1-8B & 32  & 4096  & 14336 & SiLU & ✓ \\
        Gemma-2-2B  & 26  & 2304  & 9216  & GeLUTanh & ✓ \\
        Mistral-7B  & 32  & 4096  & 14336 & SiLU & ✓ \\  
        \bottomrule
    \end{tabular}
    \label{tab:model_mlp_specs}
\end{table}

% Note that it is common that activation functions in modern LLMs allowing negative activations, such as SwiGLU \cite{shazeer2020gluvariantsimprovetransformer} in Llama 3 and GEGLU \cite{shazeer2020gluvariantsimprovetransformer} in Gemma. These negative values in the functions enable non-linear, smooth thresholding  to help models capture non-linear relationships in data.

% \section{More details on DPO algorithm}
% For an autoregressive language model (e.g., GPT-like transformers), the probability of a response $y$ given a prompt $x$ is computed as:
% \begin{equation}
%     \pi_{\theta}(y \mid x) = \prod_{t=1}^{T} \pi_{\theta}(y_t \mid x, y_{1:t-1})
% \end{equation}
% where $y_t$ is the token at position $t$ and $T$ is the response length.

% The log probability of a response is:
% \begin{equation}
%     \log \pi_{\theta}(y \mid x) = \sum_{t=1}^{T} \log \pi_{\theta}(y_t \mid x, y_{1:t-1})
% \end{equation}

% Thus, the preference log-ratio in DPO simplifies to:
% \begin{equation}
%     r_{\theta}(x, y) = \sum_{t=1}^{T} \left(\log \pi_{\theta}(y_t \mid x, y_{1:t-1}) - \log \pi_{\text{ref}}(y_t \mid x, y_{1:t-1}) \right)
% \end{equation}
% which measures how much the fine-tuned model deviates from the reference model.

\section{DPO training hyperparameters}
\label{appendix:dpo_hyperparameters}
In this section, we provide the hyperparameters for DPO training (Section~\ref{subsec:data_and_mdoels}). 

Table \ref{tab:dpo_hyperparams} reports the shared hyperparameters across models. 
Table~\ref{tab:lambda} reports the KL regularisation weight $\lambda$ tuned in DPO to maintain pre-trained model's perplexity and F1 scores for each model.

\begin{table}[h]
    \centering
    \renewcommand{\arraystretch}{1.2} % Increase row spacing
    \small
    \caption{\textit{Shared hyperparameters for DPO Training.}}
    \begin{tabular}{ll}  
        \toprule
        \textbf{Hyperparameter} & \textbf{Value / Description} \\
        \midrule
        % \textbf{Training Type} & Direct Preference Optimization (DPO) \\
        Beta ($\beta$) & 0.1 (preference strength) \\
        % \textbf{KL Weight ($\lambda$)} & 0.02 (regularisation term) \\
        Optimizer & RMSprop\\
        Learning rate & $1 \times 10^{-5}$ \\
        Warmup steps & 150 \\
        Gradient accumulation steps & 4 \\
        Batch size & 4 (per step) \\
        Evaluation batch size & 8 \\
        Max input length & 256 tokens \\
        Max new tokens & 64 tokens \\
        Max prompt length & 64 tokens \\
        Epochs & 5 \\
        % \textbf{Trainer} & BasicTrainer \\
        Gradient clipping & Max norm = 10.0 \\
        Patience for early stopping & 30 validations \\
        \bottomrule
    \end{tabular}
    \label{tab:dpo_hyperparams}
\end{table}

\begin{table}[ht]
    \centering
    \renewcommand{\arraystretch}{1.2} % Increase row spacing
    \small
    \caption{\textit{The KL regularisation weight $\lambda$ for each model}. $\lambda$ is selected to maintain perplexity and F1 scores to pre-trained models.}
    % , as it limits deviations from the pre-trained weights
    \begin{tabular}{@{}l c c@{}}
        \toprule
        \textbf{Model} & \textbf{KL weight ($\lambda$)} \\ 
        \midrule
        GPT-2-355M  & 0.02  \\ 
        Llama-3.1-8B & 0.1 \\  
        Gemma-2-2B  & 0.05  \\
        Mistral-7B  &  0.05 \\  
        \bottomrule
    \end{tabular}
    \label{tab:lambda}
\end{table}

\section{More results on toxic probes}
% validating 
\label{appendx:probe}
In this section, we provide more results on validating toxic linear probes (Section~\ref{subsec:per_neuron_toxicity_contributions}).

Table~\ref{tab:validation_accuracy} reports the test accuracies of linear probes on the Jigsaw Toxic Comment Classification dataset (90–10 split) \cite{jigsaw-toxic-comment-classification-challenge}, with all probes achieving over 91\% accuracy.
It also reports the selected 
\( \alpha \) values for probe-based steering that best preserve the pre-trained models' perplexity and F1 scores.

\begin{table}[ht]
    \centering
    \caption{\textit{Validation accuracy of toxicity probes and scaling values \( \alpha \) for probe-based steering.}  
    \( \alpha \) is selected to preserve the pre-trained  perplexity and F1 scores.}

    \renewcommand{\arraystretch}{1.2} % Increase row spacing
    \small
    \begin{tabular}{@{}l c c@{}}
        \toprule
        \textbf{Model} & \textbf{Validation Accuracy} & \textbf{\( \alpha \)} \\ 
        \midrule
        GPT-2-355M  & 95.6\% & 30  \\ 
        Llama-3.1-8B & 92.6\% & 2  \\
        Gemma-2-2B  & 96.1\% & 3  \\
        % Gemma-2-9B  & 96.1\% & 1  \\
        % Llama-2-7B & 96.4\% &   \\  
        Mistral-7B  & 91.0\%      & 5 \\  
        \bottomrule
    \end{tabular}
    \label{tab:validation_accuracy}
\end{table}

Table~\ref{tab:subtraction_more_alpha} shows that in probe-based activation steering, increasing 
\( \alpha \) beyond the selected values further reduces toxicity, but also increases perplexity and lowers F1 scores. This demonstrates a trade-off in steering: stronger steering reduces toxicity at the cost of general language quality.

\begin{table}[ht]
    \centering
    \caption{\textit{Toxicity (Toxic), log perplexity (logPPL), and F1 scores after probe-based steering with different \( \alpha \) values.}  
    Larger \( \alpha \) reduces toxicity but increases perplexity and lowers F1 scores. \textit{Bold} highlights the selected \( \alpha \) values. }

    \renewcommand{\arraystretch}{1.3} 
    \small
    \begin{tabular}{@{}p{1.8cm} p{2.3cm} p{0.7cm} p{0.7cm} p{0.7cm}@{}}
        \toprule
        \textbf{Model} & \textbf{Method} & \textbf{Toxic} & \textbf{logPPL} & \textbf{F1} \\ 
        \midrule
        GPT-2-355M & None    & 0.545 & 3.08 & 0.193 \\  
                      & Subtract (\(\alpha\)=\textbf{30}) & 0.310 & 3.19 & 0.191 \\  
                      & Subtract (\(\alpha \)=40) & 0.250 & 3.34 & 0.180 \\  
        \midrule
        Llama-3.1-8B & None     & 0.496 & 1.94 & 0.225 \\  
                      & Subtract (\( \alpha \)=\textbf{2}) & 0.335 & 2.72 & 0.187 \\ 
                      & Subtract (\( \alpha \)=3) & 0.267 & 3.53 & 0.180 \\  
        \midrule
        Gemma-2-2B   & None     & 0.488 & 4.61 & 0.231 \\  
                      & Subtract (\( \alpha \)=\textbf{3}) & 0.260 & 5.52 & 0.228 \\  
                      & Subtract (\( \alpha \)=5) & 0.251 & 5.64 & 0.226 \\  
       \midrule
        Mistral-7B   & None     & 0.507 & 1.76 & 0.231 \\  
                      & Subtract (\( \alpha \)=\textbf{5}) & 0.350 & 2.23 & 0.220 \\  
                      & Subtract (\( \alpha \)=7) & 0.319 & 2.63 & 0.212 \\  
        \bottomrule
    \end{tabular}
    \label{tab:subtraction_more_alpha}
\end{table}

\section{Negatively activated value vectors}
\label{appendix:neg_value_vectors}
In this section, 
we show that a large proportion of value vectors $v_i$ are negatively activated by their activations $m_i$ (Section~\ref{subsec:neuron_groups}).

Table~\ref{tab:percentages_neurons_neg_act} reports the percentage of MLP neurons that are negatively activated across models, showing that they constitute at least half of all MLP neurons.
% In the three larger models, approximately 50\% of MLP neurons have negative activations, while GPT-2-Medium has over 87\% of neurons negatively activated.

\begin{table}[ht]
    \centering
    \renewcommand{\arraystretch}{1.2}
    \footnotesize
     \caption{\textit{Percentages of MLP neurons with negative pre-trained activations.} 
     The three larger LLMs have approximately 50\% of their MLP neurons negatively activated, whereas GPT-2 Medium has over 87\%.}
    
    \begin{tabularx}{0.85\linewidth}{@{}l *{2}{>{\centering\arraybackslash}X}@{}}
        \toprule
        \textbf{Model} &
        \makecell{\textbf{\% neurons} \\ \textbf{negatively} \\ \textbf{activated}} &
        \makecell{\textbf{\% neurons} \\ \textbf{positively} \\ \textbf{activated}} \\
        \midrule
        GPT-2-355M  & 87.28\% & 12.71\% \\
        Llama-3.1-8B & 49.96\% & 50.04\% \\
        Gemma-2-2B  & 49.94\% & 50.06\% \\
        Mistral-7B  & 50.03\% & 49.97\% \\
        \bottomrule
    \end{tabularx}
    \label{tab:percentages_neurons_neg_act}
\end{table}

\begin{figure}[ht]
    \centering 
    \includegraphics[ width=0.475\textwidth]{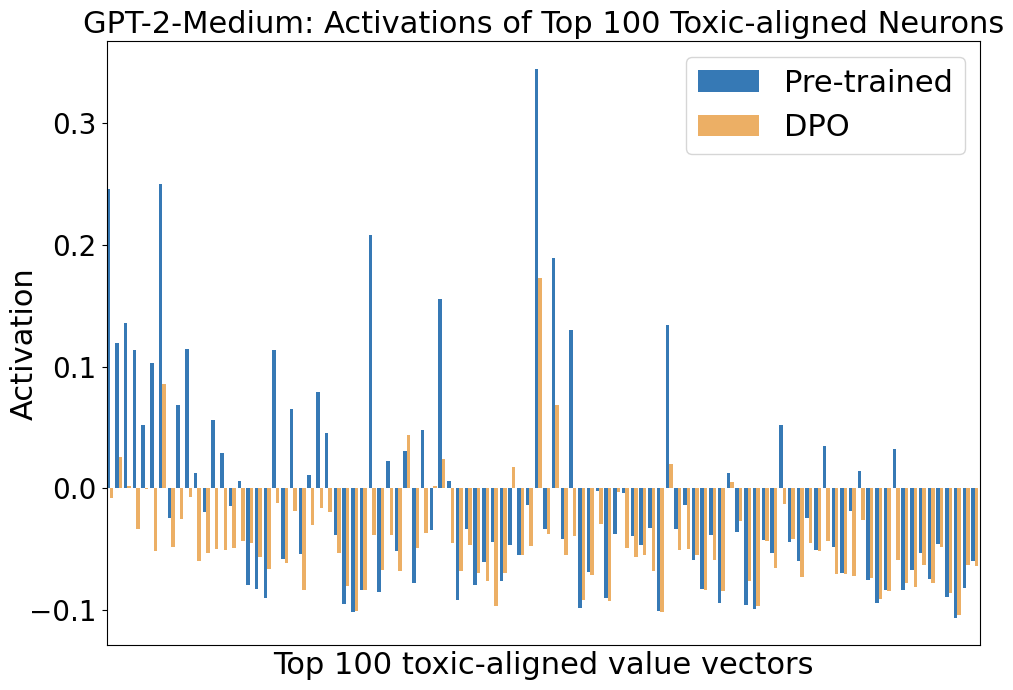}
        \caption{\textit{Activations of the top 100 toxic-aligned neurons in GPT-2-Medium.}
        The activation \( m_i \) for each value vector is averaged over all prompts and 20 generated tokens. The majority of value vectors remain weakly negatively activated both before and after DPO.}
        \label{fig:act_top_toxic_vectors}
\end{figure}

Since GPT-2 Medium has a particularly high proportion of negatively activated neurons (over 87\%), Figure~\ref{fig:act_top_toxic_vectors} illustrates this by showing the average activations of the top 100 toxic-aligned neurons. Most of these value vectors remain negatively activated both before and after DPO, reflecting the impact of the GeLU activation function.

% Table \ref{tab:disable_128_value_vectors} shows that adding more value vectors did not expand the space. The results were similar to patching and ablating $N=60$ toxic neurons. Additionally, we ablated 36 positively activated toxic neurons out of the 128, since most toxic neurons in GPT-2 medium have negative activations due to the GeLU activation function (Figure \ref{fig:act_top_toxic_vectors}). Indiscriminately ablating all toxic neurons can unintentionally increase toxicity.

\section{Logit lens tokens for value vectors}
\label{appendix:logit_lens_toxic_neurons}
In this section, we provides the tokens projected via Logit Lens for selected value vectors.

Table~\ref{tab:logit_lens_toxic} shows example toxic value vectors that project to at least one toxic token among their top-10 nearest tokens (Section~\ref{sec:dpo_incomplete}).
% across model
% Table~\ref{tab:num_toxic_neurons} reports the number of those toxic value vectors identified in each model.

Table~\ref{tab:logit_lens_anti_toxic_appendix} shows example anti-toxic value vectors that, when sign-reversed, project to at least one toxic token among their top-10 nearest tokens (Section~\ref{subsec:neuron_groups}). 
% This illustrates that, geometrically, some anti-toxic value vectors lie on the antipode of toxic semantic clusters.

% For GPT2, this yields $N=60$ neurons, as increasing the threshold beyond this point results in value vectors no longer projecting onto toxic tokens using the Logit Lens. 
% We also test a looser threshold of $N=128$, where the toxic subspace formed by the top 128 value vectors in GPT-2-Medium stabilises under singular value decomposition (SVD), with additional value vectors failing to expand the space.

% \citet{lee2024mechanisticunderstandingalignmentalgorithms} supported their claim by amplifying the top 7 toxic neurons' activations in the DPO-ed model, scaling their key vectors by a factor of 10, and reversed the toxicity (Table \ref{tab:disable_value_vectors}). However, we argue this intervention does not causally prove that dampening these neurons is DPO's primary mechanism. Amplifying these neurons by 10x drastically increases their impact beyond pre-DPO levels, likely to raise toxicity by boosting the norm of the toxic direction in the residual stream, similar to adding a steering vector \cite{tan2024analyzinggeneralizationreliabilitysteering}. This contrast echos the observation that inducing behaviour through unrealistic interventions on a model's component does not prove that this component alone is responsible for the behaviour \cite{alignmentforum2023}, as seen in activation patching.

\begin{table}[ht]
% \centering
\renewcommand{\arraystretch}{1.3}
\arrayrulecolor{black}
\small
\vspace{-0.5em}
\caption{\textit{Examples of toxic value vectors that project to toxic tokens via LogitLens.} 
Value vectors are \textit{ranked} by their cosine similarity with the toxicity probe $W_{\text{Toxic}}$.
Table~\ref{tab:num_toxic_neurons} provides the number of such toxic value vectors. 
\textcolor{red}{Warning: these examples are highly offensive.}}
\label{tab:logit_lens_toxic}

\begin{subtable}[t]{\textwidth}
\hspace{30mm}
\noindent GPT-2 medium
\\
% \vspace{5mm}
\begin{tabular}{@{}p{1.1cm} p{0.6cm} p{5.1cm}@{}}
\toprule
\textbf{Vector} & \textbf{Rank} & \textbf{Top tokens} \\
\midrule
$\mathbf{W_{\text{Toxic}}}$ & - & c*nt, f*ck, a**hole, d*ck, wh*re, holes \\
$\mathbf{v_{19}^{770}}$ & 1 & sh*t, a**, cr*p, f*ck, c*nt, garbage \\ % trash 
% $\mathbf{v_{18}^{2669}}$ & 3 & degenerate, whining, idiots, stupid \\ % smug
% MLP.$\mathbf{v_{3}^{3680}}$ & 10 & se*ist, feminist, Femin, femin, misogyn \\
$\mathbf{v_{13}^{253}}$ & 18 & c*m, d*ck, icles, icle, bo*bs, naughty \\
$\mathbf{v_{7}^{3358}}$ & 29 & cr*p, whine, sh*t, uphem, shri, bullsh*t \\
% MLP.$\mathbf{v_{15}^{1696}}$ & 44 & death, extermination, Corpse, decap \\
$\mathbf{v_{6}^{3972}}$ & 50 & f*cking, d*mn, sinful, hell, immoral \\
% MLP.$\mathbf{v_{12}^{2756}}$ & 59 & burdens, bad, offending, imped, horrible \\
% MLP.$\mathbf{v_{1}^{2057}}$ & 60 & Bench, rodu, Sequ, RIP, Brist, Vers \\
\bottomrule
\end{tabular}
\end{subtable}

\vspace{1em}

\begin{subtable}[t]{\textwidth}
\hspace{30mm}
\noindent Llama-3.1-8B

\begin{tabular}{@{}p{1.1cm} p{0.6cm} p{5.1cm}@{}}
\toprule
\textbf{Vector} & \textbf{Rank} & \textbf{Top tokens} \\
\midrule
$\mathbf{W_{\text{Toxic}}}$ & - & en, kommen, F*CK, iyah, f*ck, dirty \\ 
$\mathbf{v_{25}^{14065}}$ & 1 & sh*t, cr*p, rubbish, garbage, bullsh*t \\ % junk, nonsense, trash
% MLP.$\mathbf{v_{18}^{15439}}$ & 2 & fu*k, Fu*k, fu*ks, fu*ked, fu*king \\ %, FUCK, shit
% MLP.$\mathbf{v_{81}^{396}}$ & 3 & sh*t, fu*k, Fu*k, fu*ked, Fu*king\\
% fucks, assh 
% $\mathbf{v_{19}^{1569}}$ & 5 & cr*p, sh*t, hell, living, Hell, tar, he*k\\
% MLP.$\mathbf{v_{19}^{7649}}$ & 6 & stuff, sh*t, Stuff, fu*k, guy, fu*king\\
$\mathbf{v_{27}^{3863}}$ & 7 & waste, trash, garbage, junk, wastes\\
$\mathbf{v_{18}^{13797}}$ & 10 & Shoot, oh, sh*t, cr*p, swore, allocator\\
$\mathbf{v_{19}^{6041}}$ & 18 & d*mn, sh*t, DAM, dam, fig, D*mn\\
\bottomrule
\end{tabular}
\end{subtable}

\vspace{1em}

\begin{subtable}[t]{\textwidth}
\hspace{30mm}
\noindent Gemma-2-2B

\begin{tabular}{@{}p{1.1cm} p{0.6cm} p{5.1cm}@{}}
\toprule
\textbf{Vector} & \textbf{Rank} & \textbf{Top tokens} \\
\midrule
$\mathbf{W_{\text{Toxic}}}$ & - & rungsseite, fu*k, Fu*king, SH*T \\ % a**hole
$\mathbf{v_{4}^{2288}}$ & 1 & idiot, bastard, a*hole, fu*ks, moron \\ % FUCK 
$\mathbf{v_{18}^{2597}}$ & 10 & yes, ridiculous, absurd, ludicrous \\ %StatelessWidget, pathetic, 'endwhile', ' awful' 
$\mathbf{v_{3}^{4727}}$ & 22 & sh*t, Sh*t, cr*p, sh*ts, sh*te, sh*tty \\
\bottomrule
\end{tabular}
\end{subtable}

\vspace{1em}

\begin{subtable}[t]{\textwidth}
\hspace{30mm}
\noindent Mistral-7B

\begin{tabular}{@{}p{1.1cm} p{0.6cm} p{5.1cm}@{}}
\toprule
\textbf{Vector} & \textbf{Rank} & \textbf{Top tokens} \\
\midrule
$\mathbf{W_{\text{Toxic}}}$ & - & sh*t, f*ck, assh, bullsh*t, a**hole \\ %f*cked, 
$\mathbf{v_{22}^{1061}}$ & 1 & fu*k, sh*t, bullsh*t, a**hole, sh*tty\\ % assh
% MLP.$\mathbf{v_{14}^{2292}}$ & 3 & sh*t, cr*p, da*n, fu*king, bullsh*t \\ % stupid, su*ks
$\mathbf{v_{15}^{2454}}$ & 4 & fuck*ng, bullsh*t, stupid, sh*t, cr*p\\ %piss, disgust, ridiculous 
% MLP.$\mathbf{v_{22}^{5047}}$ & 8 & sh*t, ***, **, fu*king, Fu*k \\
% MLP.$\mathbf{v_{20}^{1349}}$ & 17 & ass, fool, ASS, idiot, hole \\
$\mathbf{v_{14}^{11281}}$ & 34 & sexual, sex, girls, women, dating, porn\\
$\mathbf{v_{19}^{4689}}$ & 45 & cr*p, sh*t, d*mn, hell, b*tch, piss \\
% $\mathbf{v_{6}^{7364}}$ & 48 & lie, fu*k, da, foul, bitch, cr*p, nuts \\
\bottomrule
\end{tabular}
\end{subtable}

\end{table}

\begin{table}[t!]
% \centering
\renewcommand{\arraystretch}{1.3}
\arrayrulecolor{black}
\small
\vspace{-0.5em}
\caption{Examples of anti-toxic value vectors that, when sign-reversed, project to toxic tokens via Logit Lens.  
\textit{Rank} gives the cosine similarity rank with $-1 \!\times\! W_{\text{Toxic}}$, reflecting how ``anti-toxic'' a neuron is.
\textcolor{red}{Warning: these examples are highly offensive.}}
\label{tab:logit_lens_anti_toxic_appendix}

\begin{subtable}[t]{\textwidth}
\hspace{30mm}
\noindent GPT-2 medium

\begin{tabular}{@{}p{1.55cm} p{0.45cm} p{5cm}@{}}
\toprule
\textbf{Vector} & \textbf{Rank} & \textbf{Top tokens} \\
\midrule
$-1\!\times\!\mathbf{v_{10}^{1882}}$ & 1 &  maniac, ueless, thug, arrog, f*cking \\ % disregard
$-1\!\times\!\mathbf{v_{11}^{1307}}$ & 3 & d*mn, darn, kidding, freaking, piss \\ 
$-1\!\times\!\mathbf{v_{15}^{301}}$ & 4 & harmful, worse, unfavorable, disturbing \\ % unpleasant, detrimental, undesirable, restrictive
\bottomrule
\end{tabular}
\end{subtable}

\vspace{1em}

\begin{subtable}[t]{\textwidth}
\hspace{30mm}
\noindent Llama-3.1-8B

\begin{tabular}{@{}p{1.55cm} p{0.45cm} p{5cm}@{}}
\toprule
\textbf{Vector} & \textbf{Rank} & \textbf{Top tokens} \\
\midrule
$-1\!\times\!\mathbf{v_{25}^{14671}}$ & 2 & f*ck, f*cked, f*cking, sh*t, F*CK \\ 
$-1\!\times\!\mathbf{v_{14}^{4997}}$ & 19 & s*cks, s*ck, adla, BackPressed, teri \\ %awesome
\bottomrule
\end{tabular}
\end{subtable}

\vspace{1em}

\begin{subtable}[t]{\textwidth}
\hspace{30mm}
\noindent Gemma-2-2B

\begin{tabular}{@{}p{1.55cm} p{0.45cm} p{5cm}@{}}
\toprule
\textbf{Vector} & \textbf{Rank} & \textbf{Top tokens} \\
\midrule
$-1\!\times\!\mathbf{v_{14}^{7822}}$ & 1 & f*cking, godd*mn, f*ck, sh*t, d*mn\\ 
$-1\!\times\!\mathbf{v_{6}^{7099}}$ & 2 & f*cking, f*ck, f*cker, p*ss, F*ck \\  
$-1\!\times\!\mathbf{v_{17}^{8418}}$ & 13 & idiot, idiots, stupid, moron, dumbass \\
\bottomrule
\end{tabular}
\end{subtable}

\vspace{1em}

\begin{subtable}[t]{\textwidth}
\hspace{30mm}
\noindent Mistral-7B

\begin{tabular}{@{}p{1.55cm} p{0.45cm} p{5cm}@{}}
\toprule
\textbf{Vector} & \textbf{Rank} & \textbf{Top tokens} \\
\midrule
$-1\!\times\!\mathbf{v_{14}^{14693}}$ & 1 & sh*t, f*ck, Block, piss, f*cking, bitch \\
$-1\!\times\!\mathbf{v_{14}^{8200}}$ & 16 & cr*p, nonsense, stupid, d*mn, ridiculous \\ % bullshit, shit 
$-1\!\times\!\mathbf{v_{17}^{14302}}$ & 25 & hell, d*mn, d*mned, f*ck, cr*p, sh*t \\
$-1\!\times\!\mathbf{v_{12}^{8139}}$ & 36 & f*cked, sh*t, bitch, sex, sexual, rape \\
% piss, suck
% MLP.$\mathbf{v_{23}^{11609}}$ & xx & nasty, nast, sol, rezent, erra, ket \\
\bottomrule
\end{tabular}
\end{subtable}

\end{table}

\section{Projecting value vectors to a toxic subspace}
\label{appendix:why_not_toxic_subspace}
In this section, 
we present initial results using a toxic subspace to capture toxicity representations in GPT-2-Medium and to perform projections (discussed in \textit{Limitations}).
We explain why we do not adopt this approach for neuron analysis, as it complicates the identification of coherent neuron groups.

Specifically, on GPT-2-Medium, we apply singular value decomposition (SVD) to the value vectors of 128 toxic-aligned MLP neurons, 
using the top three components as basis directions to capture different aspects of toxicity. 
We choose $N=128$ because it yields a stable toxic subspace—adding more value vectors does not significantly expand it. 
Table~\ref{tab:logit_lens_svd} shows that these SVD vectors unembed to different toxic tokens, including offensive curse words (\(\text{SVD}_{\text{Toxic}}[0]\)), mild insults (\(\text{SVD}_{\text{Toxic}}[1]\)), and sexualised terms (\(\text{SVD}_{\text{Toxic}}[2]\)).

\begin{table}[ht]
    \centering
    \renewcommand{\arraystretch}{1.2} % Adjust row spacing for better readability
    \small
    \caption{\textit{Logit Lens tokens for the top three SVD vectors extracted from 128 toxic-aligned neurons in GPT-2 Medium.}  
    Each SVD direction captures a different aspect of toxicity. 
    \textcolor{red}{Warning: these examples are highly offensive.}}
    \label{tab:logit_lens_svd}
    
    \begin{tabularx}{0.48\textwidth}{@{}l X@{}} % Use tabularx to control column width
        \toprule
        \textbf{Model} & \textbf{Top Tokens} \\ 
        \midrule
        \(\text{SVD}_{\text{Toxic}}[0]\) & f*ck, assh*le, f*cking, d*ck, sh*t, sl*t \\ 
        \(\text{SVD}_{\text{Toxic}}[1]\) & d*mned, cr*p, stupid, darn, Godd, idiots\\ 
        \(\text{SVD}_{\text{Toxic}}[2]\) & sex, boobs, chicks, sexy, vagina, breasts\\
        \bottomrule
    \end{tabularx}
\end{table}

Follow Section~\ref{subsec:neuron_groups}, we attempt to identify neuron groups based on their projection changes onto the toxicity subspace.
One approach is to compute a weighted sum of the SVD vectors (scaled by their singular values) to form a single combined direction, then measure projections onto it.
However, this provides little advantage over using a standard toxicity probe. 
Instead, we project each value vector onto each  SVD vectors individually.

Since the SVD vectors are orthonormal, the total projection onto the toxic subspace is equivalent to summing the projections onto each SVD direction.
Thus to identify neurons reducing toxicity, we compute each value vector’s cosine similarity with the SVD vectors, along with their projections before and after DPO. 

We find that 74.7\% of value vectors have conflicting signs of alignment across the SVD directions—that is, they align positively with at least one vector and negatively with another.
This complicates defining whether a neuron is ``toxic-aligned''. 
Similarly, 74.3\% of neurons show inconsistent projection change after DPO, reducing toxicity along one direction while increasing it along another.
% Only 22.6\% of neurons have consistent signs across all three directions for both alignment and projection change.

These inconsistencies make it impossible to identify coherent neuron groups that reduce toxicity across all SVD directions, i.e. across the toxic subspace. 
This also means that each SVD direction induces its own set of contradictory neuron groups. 
% making group-based attribution unreliable.
More importantly, this prevents us from linking toxicity scores to specific neuron groups via activation patching (Section~\ref{subsec:neuron_groups}), 
as a single neuron can simultaneously increase and decrease toxicity depending on the direction. 
% Assigning neurons to the SVD direction with the largest projection change oversimplifies each neuron's role and undermines interpretability.

For these reasons, we choose not to proceed with subspace projection for neuron analysis and instead focus on the single-probe approach.

\section{More results on activation shifts}
\label{appendix:activation_shifts}
In this section, we provide more results on DPO-induced activation shifts by presenting their distributions and analyse whether they occur systematically  with neuron properties.
These results complement Section~\ref{subsec:balance_opposing_effects}.

\begin{figure*}[ht]
        \centering
\includegraphics[width=0.95\textwidth]{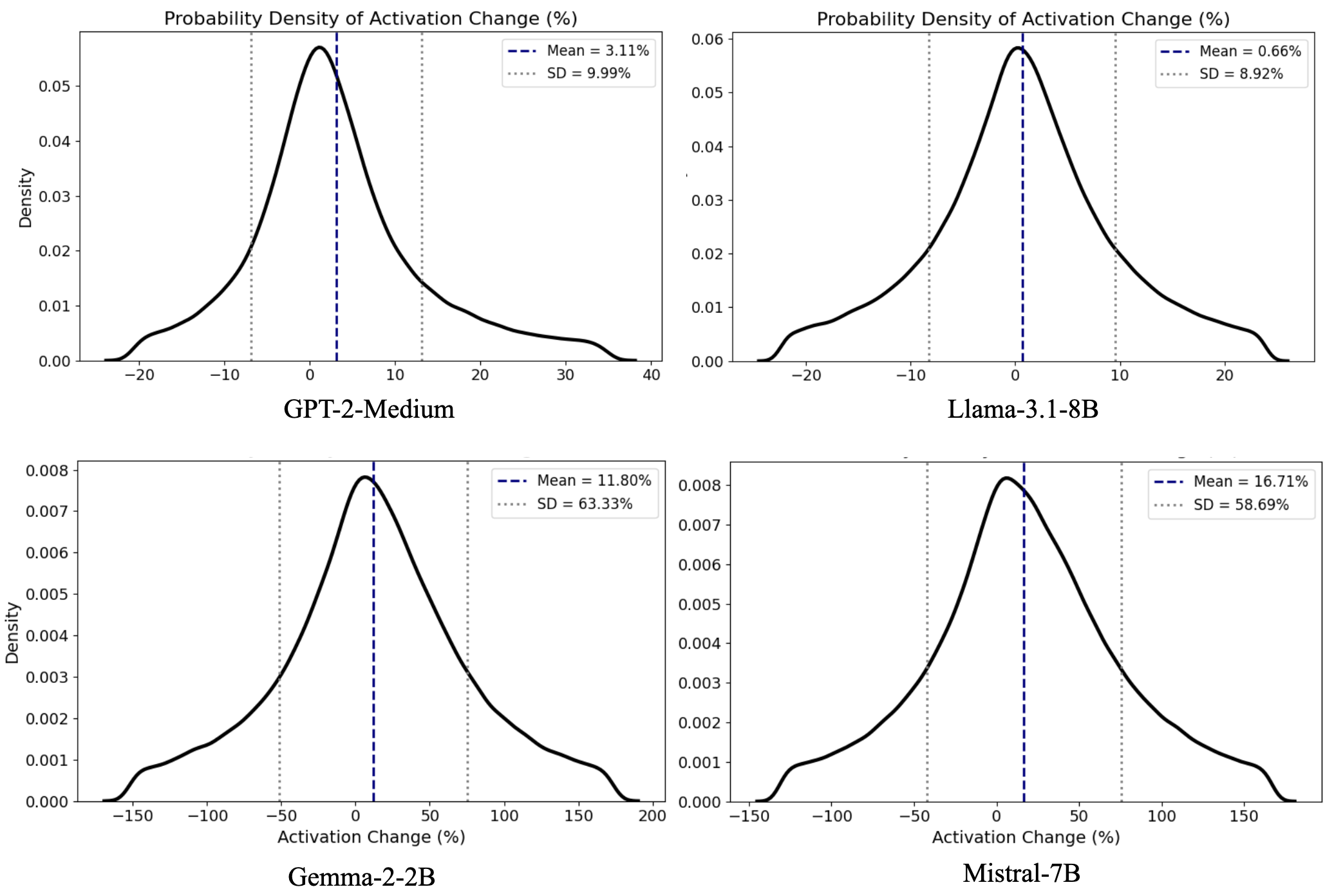}
    \caption{\textit{Probability density of activation shifts ($m_i^{\text{pre}}-m_i^{\text{dpo}}$) during DPO.} 
    Most neurons have small activation shifts around the mean, with more substantial variation in the tails. 
    Gemma-2-2B and Mistral-7B show larger average shifts and standard deviations (SD) compared to the other two models. }
    % Activation shifts are clipped to the [10\%, 90\%] quantile range to show the main distributions.
\label{fig:four_model_act_dist}
\end{figure*}

Figure~\ref{fig:four_model_act_dist} shows the distribution of activation shifts across models. 
Most neurons have small activation shifts around the mean but substantial variation in the tails.
% the distribution sub-figure for each group
% Further breaking down the activation shifts for each neuron group shows that, \TP{} and \AN{} have an average positive shift, corresponding to reducing toxicity writing; \TN{} and \AP{} have an average negative shift, corresponding to promoting anti-toxicity. 
% These patterns align with the activation movements of the four groups during DPO (see Figure~\ref{fig:Llama3_illustration_activation_shifts}). 

Table~\ref{tab:act_correlation_analysis} presents the results of a Pearson correlation analysis \cite{schober2018correlation} between DPO-induced activation shifts and neuron properties. 
The analysis reveals no correlation between activation shifts and the ``toxicity level'' of a neuron—measured by its cosine similarity with the toxic probe—and only a weak positive correlation with pre-trained activations. While this may suggest a slight tendency for DPO to push activations toward zero, 
the pattern is likely due to a regression-to-the-mean effect, thus more of a statistical artifact than an intentional toxicity-reduction mechanism. These findings indicate that DPO-induced activation shifts are largely random.
% neurons with higher positive activations tend to undergo larger decreases, while highly negative activations tend to increase during DPO, pushing activations toward zero.

% \begin{figure*}[ht]
%     \centering
%     \begin{subfigure}[b]{0.45\textwidth}
%         \centering
%         \includegraphics[width=\textwidth]{Figures/gpt2_act_dist_by_group.png}
%         \caption{GPT-2-355M}
%     \end{subfigure}
%     \begin{subfigure}[b]{0.45\textwidth}
%         \centering
%         \includegraphics[width=\textwidth]{Figures/Llama3_act_dist_by_group.png}
%         \caption{Llama-3.1-8B}
%     \end{subfigure}
%     \begin{subfigure}[b]{0.45\textwidth}
%         \centering
%         \includegraphics[width=\textwidth]{Figures/gemma2_act_dist_by_group.png}
%         \caption{Gemma-2-2B}
%     \end{subfigure}
%     \begin{subfigure}[b]{0.45\textwidth}
%         \centering\includegraphics[width=\textwidth]{Figures/mistral_act_dist_by_group.png}
%         \caption{Mistral-7B}
%     \end{subfigure}
%     \caption{\textit{Probability density of activation shifts during DPO across four groups.} 
%     \TP{} and \AN{} show an average positive shift, corresponding to reducing toxicity writing; \TN{} and \AP{} show an average negative shift, corresponding to promoting anti-toxicity.  
%     Activation shifts are clipped to the [10\%, 90\%] quantile range to show the main distributions.}
% \label{fig:four_model_act_four_group}
% \end{figure*}

\begin{table*}[ht]
    \centering
    \renewcommand{\arraystretch}{1.2}
    \footnotesize
    \caption{\textit{Pearson correlation between activation shifts and neuron properties.}  
Activation shifts ($m_i^{\text{pre}} - m_i^{\text{dpo}}$) show no correlation with a neuron's "toxicity level" (measured by cosine similarity with the toxic probe), and only a weak positive correlation with pre-trained activations, which is likely a regression-to-the-mean effect.
}
% This slight tendency for DPO to push activations toward zero may simply reflect a regression-to-the-mean effect.
    
    \begin{tabularx}{\linewidth}{@{}l l *{4}{>{\centering\arraybackslash}X}@{}}
        \toprule
        \textbf{Variables} & \textbf{Metric} & \textbf{GPT-2-355M} & \textbf{Llama-3.1-8B} & \textbf{Gemma-2-2B} & \textbf{Mistral-7B} \\
        \midrule
        \multirow{2}{*}{\makecell[l]{Activation shift \\\& probe alignment}} 
            & Correlation & 0.004 & 0.001 & 0.004 & 0.003 \\
            & p-value             & 0.252 & 0.487 & 0.071 & 0.045 \\
        \midrule
        \multirow{2}{*}{\makecell[l]{Activation shift \\\& pre-trained activation}} 
            & Correlation & 0.263 & 0.033 & 0.098 & 0.347 \\
            & p-value             & \textbf{<0.0001} & \textbf{<0.0001} & \textbf{<0.0001} & \textbf{<0.0001} \\
        \bottomrule
    \end{tabularx}
    \label{tab:act_correlation_analysis}
\end{table*}

% \begin{figure*}[t!]
%     \centering
%     \begin{subfigure}[b]{0.45\textwidth}
%         \centering
%         \includegraphics[width=\textwidth]{Figures/gpt2_act_random.png}
%         \caption{GPT-2-355M}
%     \end{subfigure}
%     \begin{subfigure}[b]{0.45\textwidth}
%         \centering
%         \includegraphics[width=\textwidth]{Figures/Llama3_act_random.png}
%         \caption{Llama-3.1-8B}
%     \end{subfigure}
%     \begin{subfigure}[b]{0.45\textwidth}
%         \centering
%         \includegraphics[width=\textwidth]{Figures/gemma2_act_random.png}
%         \caption{Gemma-2-2B}
%     \end{subfigure}
%     \begin{subfigure}[b]{0.45\textwidth}
%         \centering\includegraphics[width=\textwidth]{Figures/mistral_act_random.png}
%         \caption{Mistral-7B}
%     \end{subfigure}
%     \caption{\textit{Clusters of neuron activation change vs. probe alignment across four models.} 
%     Each point represents an MLP neuron, plotted by its cosine similarity with the toxicity probe (x-axis) and its change in activation ($m_i^{pre} - m_i^{dpo}$ after DPO (y-axis). 
%     Across all models, activation shifts appear randomly distributed: neurons with similar cosine similarities show both positive and negative activation changes without a consistent pattern.}
% \label{fig:four_model_random_act_diff}
% \end{figure*}

\section{More results on opposing neuron effects}
\label{appendix:opposing_neuron_effects}
In this section, we provide more statistics and visualisations on the opposing neuron effects (Section~\ref{subsec:balance_opposing_effects}). 

Table~\ref{tab:percentages_increase_decrease} shows the percentage of neurons reducing toxicity projection ($\Delta_{\text{Toxic}, i} < 0$, denoted as $\downarrow$), 
ranging from 52\% in Gemma-2-2B to 58\% in GPT-2-Medium.
This shows that DPO's activation shifts cause roughly half of the MLP neurons to reduce toxicity projection, while the other half increase it, revealing a trade-off in toxicity reduction.

\begin{figure*}[t!]
    \centering \includegraphics[width=0.96\textwidth]{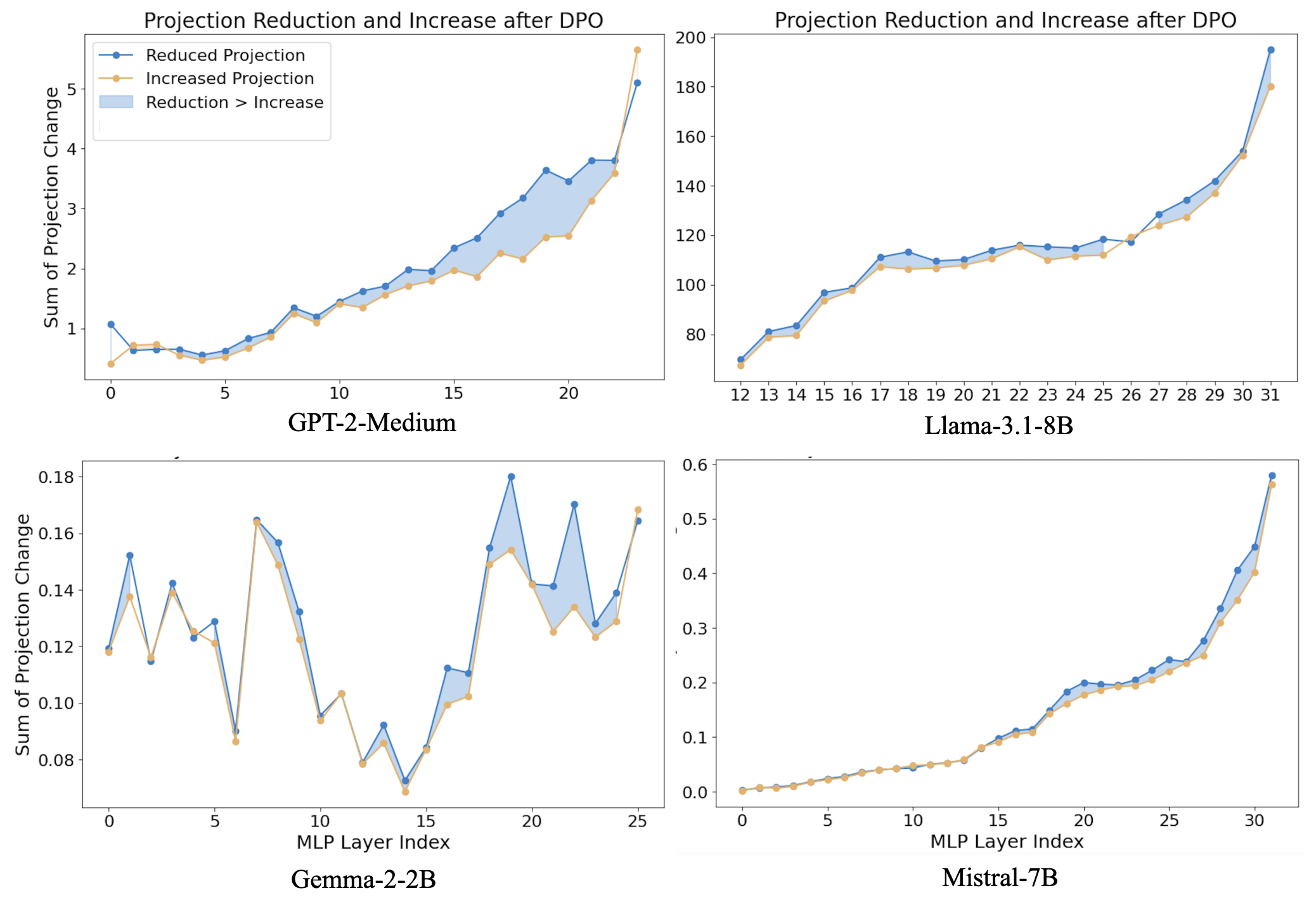}
   \caption{\textit{  
   DPO balances opposing toxicity writing across \textit{all} MLP layers.} 
   % through distributed activation shifts 
   Blue dots show the total projection reduction per layer, 
   while orange dots show the total increase, both after DPO. 
   The shaded blue areas illustrate how the opposing effects cancel out and lead to a net toxicity reduction. 
   Projection changes tend to grow in later layers when measured against the last-layer probe.}
   % though this trend is less evident for Gemma-2-2B
   \label{fig:four_model_net_reduction_full}
\end{figure*}

Figure~\ref{fig:four_model_net_reduction_full} visualises the opposing effects across all MLP layers, complementing Figure~\ref{fig:four_model_net_reduction} by including the first 10 layers that were omitted.

\begin{table}[ht]
    \centering
    \renewcommand{\arraystretch}{1.2}
    \footnotesize
    \caption{\textit{Percentages of neurons reducing toxicity projection after DPO.}  
    Across models, 52\% to 58\% of MLP neurons reduce their projection ($\Delta_{\text{Toxic}, i} < 0$) onto the toxicity probe, while the remaining neurons increase it ($\Delta_{\text{Toxic}, i} > 0$).}

    \resizebox{1\linewidth}{!}{ % Reduce width to 90% of text width
    \begin{tabular}{@{}p{1.8cm}cc@{}}
        \toprule
        \textbf{Model} & 
        \makecell{\textbf{\% neurons} \\ \textbf{reduce projection (\(\downarrow\))}} &
    \makecell{\textbf{\% neurons} \\ \textbf{increase projection (\(\uparrow\))}} \\
        \midrule
        GPT-2-355M    & 58.49\% & 41.51\% \\
        Llama-3.1-8B  & 53.01\% & 46.99\% \\
        Gemma-2-2B    & 51.75\% & 48.25\% \\
        Mistral-7B    & 51.98\% & 48.02\% \\
        \bottomrule
    \end{tabular}
    }
    \label{tab:percentages_increase_decrease}
\end{table}

% Table \ref{tab:percentages_full} computes three quantities:
% (a) the net toxicity reduction, as the sum of all per-neuron toxicity changes, following Equation \ref{eq:toxicity_reduction}:
% {\fontsize{10}{12}\selectfont
% \[
% \text{net\_reduction} = \sum_{i} \Delta_{\text{Toxic}, i}
% \]
% }
% \noindent
% (b) The total toxicity reduction, representing the sum of all positive toxicity changes:
% {\fontsize{10}{12}\selectfont
% \[
% \text{total\_reduction} = \sum_{i:\Delta_{\text{Toxic}, i} > 0} \Delta_{\text{Toxic}, i}
% \]
% }
% \noindent
% (c) The proportion of net toxicity reduction relative to the total reduction from neurons with  
% positive toxicity changes:
% {\fontsize{10}{12}\selectfont
% \[
% \text{relative\_reduction} = \frac{\text{net\_reduction}}{\text{total\_reduction}}
% \]
% }

% Table \ref{tab:relative_reduction_ratios} shows that across four models, the net reduction accounts for 4\% to 15\% of the maximum toxicity reduction induced by DPO. This suggests that most reductions are counteracted by neurons increasing toxicity, leaving only a small net decrease after DPO. Note that these percentages do not account for neuron interactions across layers.

\section{More results on four neuron groups}
\label{appendix:four_neuron_groups}
In this section, we provide more visualisations on the four neuron groups (Section~\ref{subsec:neuron_groups}). 

% We repeat the analysis for Figure~\ref{fig:neuron_group_contributions} on the three other models (excluding Llama-3.1-8B), and present the per-layer toxicity reduction for each group across models.
% Finally, we report the statistics on the four broad groups—\TPbroad{}, \TNbroad{}, \APbroad{}, and \ANbroad{} (defined in Section~\ref{subsec:replicating_dpo_with_activation_editing}) 

\begin{figure*}[ht]
    \centering

    % ====== First Model ======
    \begin{subfigure}[t]{\textwidth}
\includegraphics[width=0.94\textwidth]{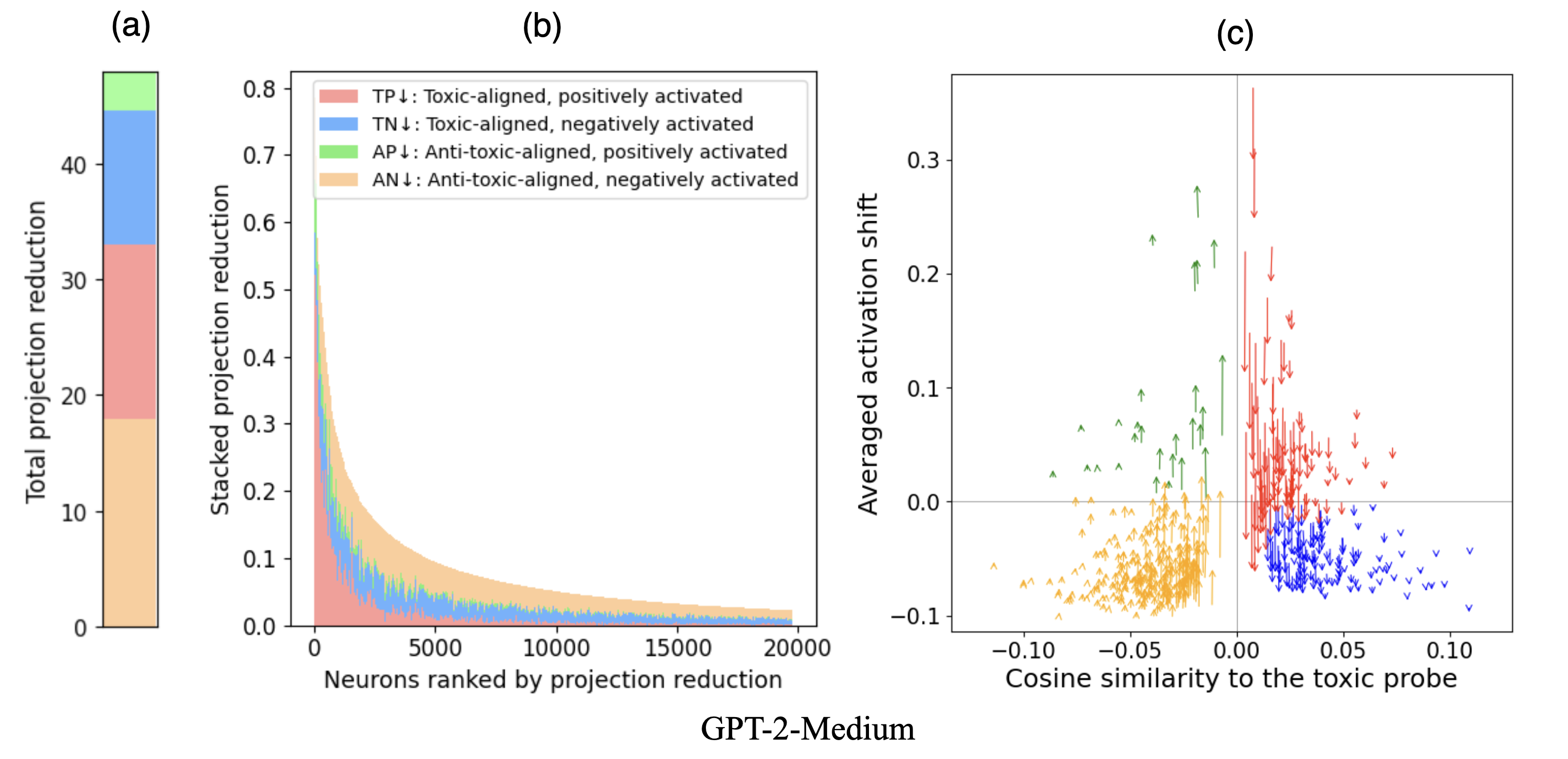}
        \caption*{}
    \end{subfigure}
    \vspace{-6mm} % reduce vertical gap

    % ====== Second Model ======
    \begin{subfigure}[t]{\textwidth}
\includegraphics[width=0.94\textwidth]{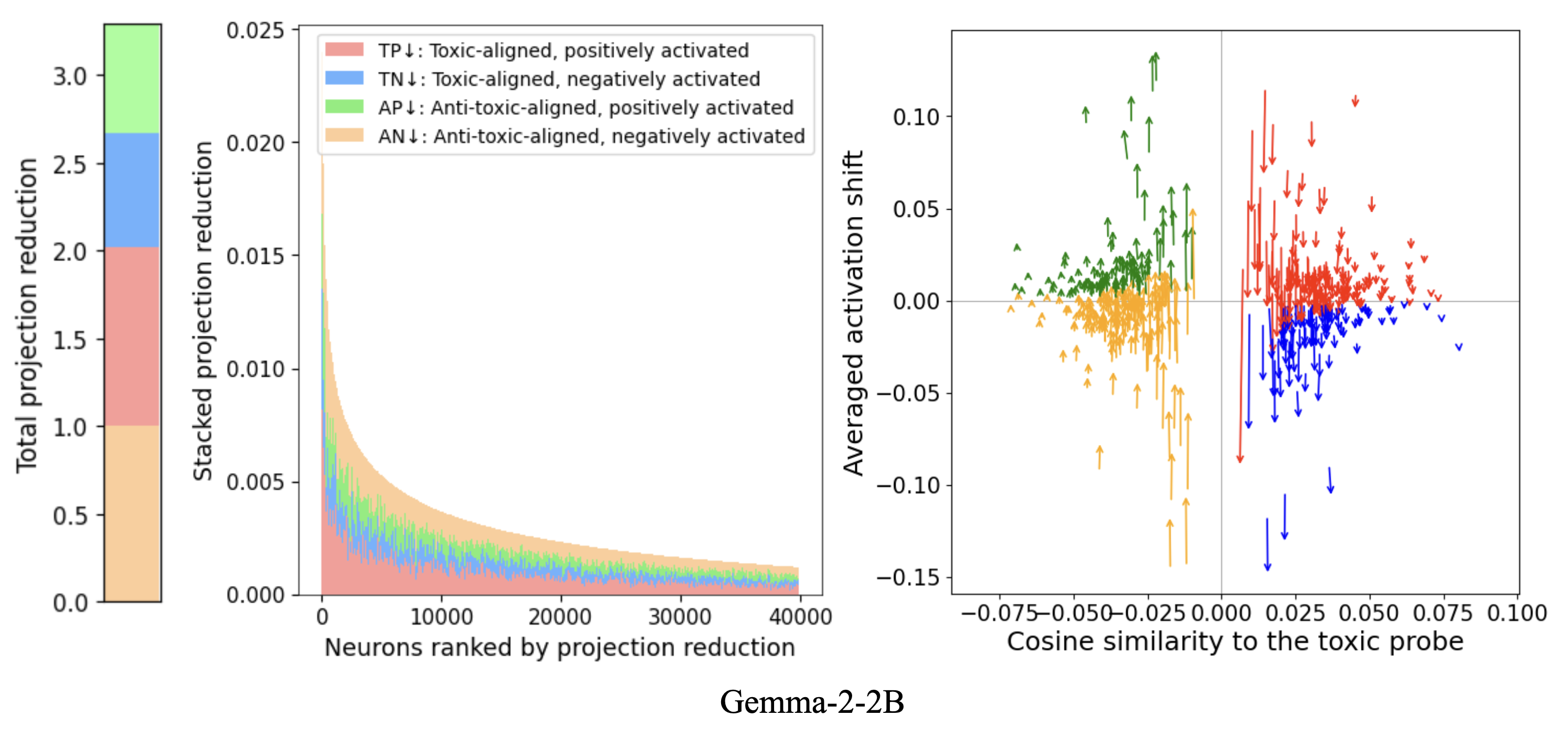}
        \caption*{}
    \end{subfigure}
    \vspace{-6mm} % reduce vertical gap

    % ====== Third Model ======
    \begin{subfigure}[t]{\textwidth}
\includegraphics[width=0.94\textwidth]{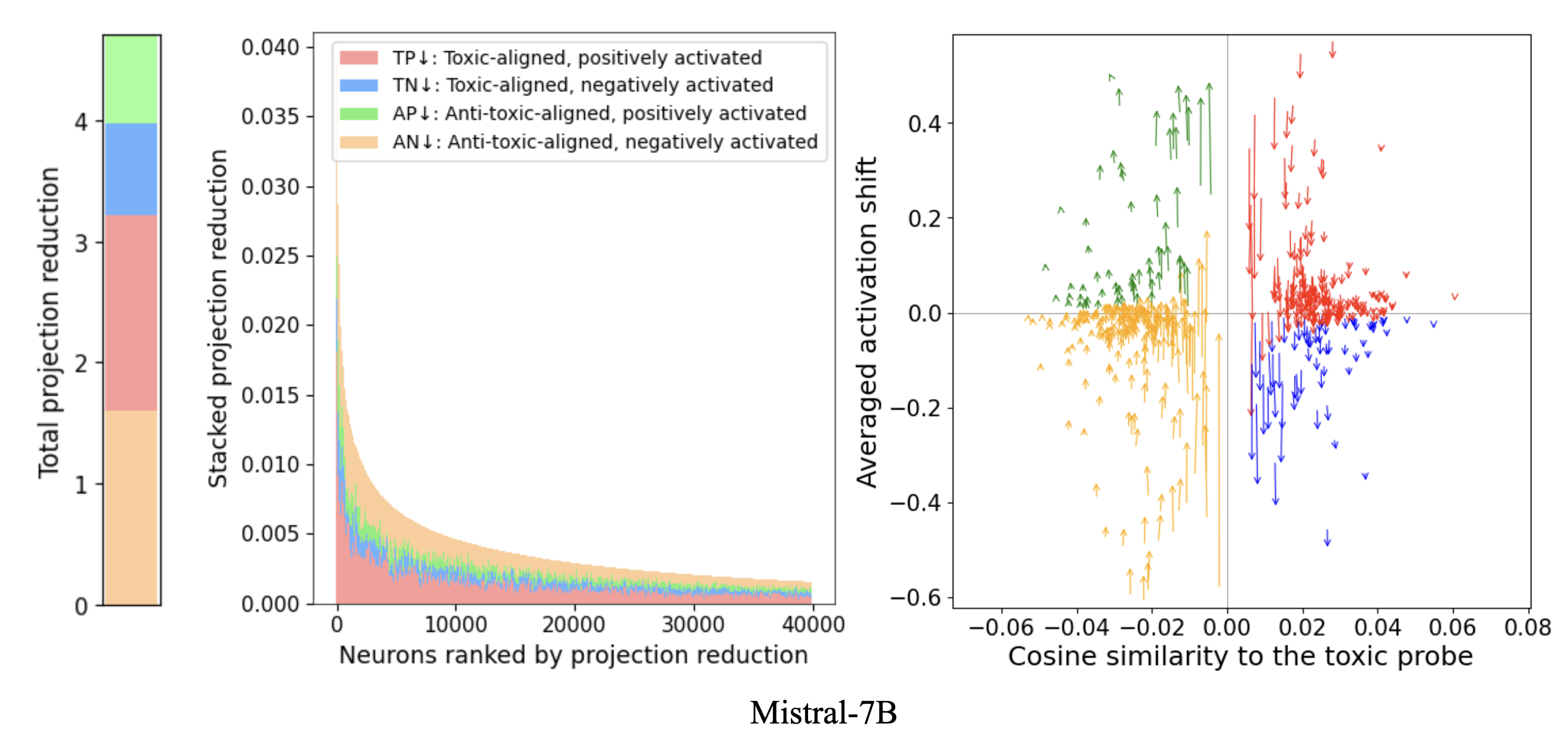}
        \caption*{}
    \end{subfigure}
    \vspace{-5mm} % reduce space before caption
    \caption{
    \textit{Four neuron groups collectively reduce toxicity during DPO, shown for GPT-2-Medium, Gemma-2-2B, and Mistral-7B.} 
    The same four groups consistently emerge as in Llama-3.1-8B. 
    % For each model, the three vertical panels show:  
    (a) Proportion of toxicity reduction per group, where \TP{} and \AN{} dominate;  
    (b) Cumulative toxicity reduction for the top 40,000 neurons (ranked by reduction in projection), where \TP{} dominates the early ranks and \AN{} gradually catches up;  
    (c) Per-group activation shifts during DPO for the top 2,000–2,500 neurons, where each group shifts according to its orientation relative to the toxic representation.}
    \label{fig:neuron_group_contributions_all_models}
\end{figure*}

Figure~\ref{fig:neuron_group_contributions_all_models} shows the four-group distributions for GPT-2-Medium, Gemma-2-2B, and Mistral-7B, repeating the analysis from Figure~\ref{fig:neuron_group_contributions} for Llama-3.1-8B.
In these three models, overall toxicity reduction is primarily driven by \textcolor{red}{$\rm TP\downarrow$} and \textcolor{orange}{$\rm AN\downarrow$}, which dominate the stacked bars in Figure~\ref{fig:neuron_group_contributions_all_models}a.

\begin{figure*}[ht]
\centering
\includegraphics[width=0.9\linewidth]{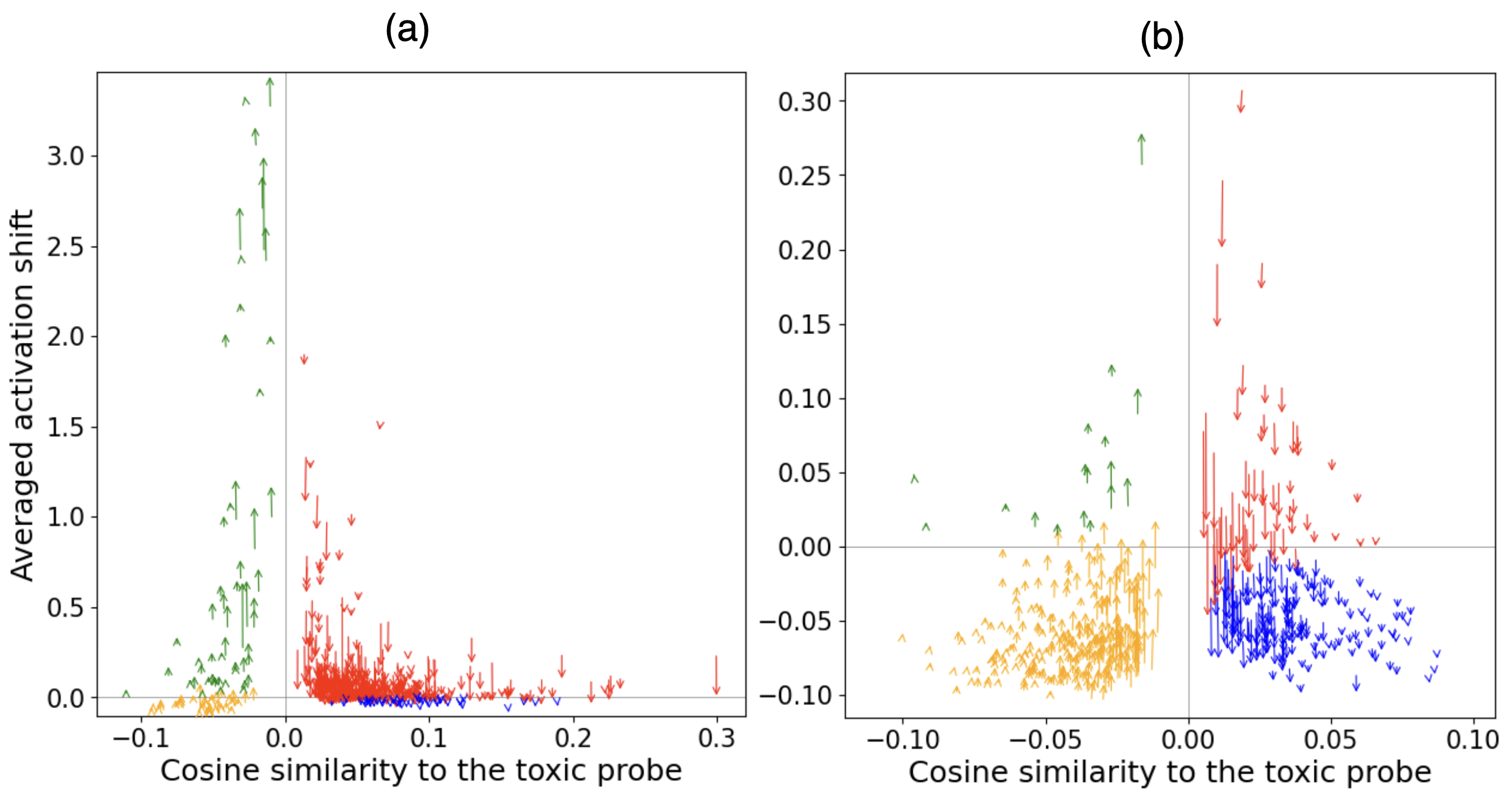}
\caption{\textit{Activation shifts of top-contributing neurons to toxicity projection reduction in GPT-2-Medium.}
(a) Activation shifts of top 500 neurons, where \textcolor{red}{$\rm TP\downarrow$} drives the reduction.
(b) Activation shifts of neurons ranked 5000–5500, 
showing increased \textcolor{orange}{$\rm AN\downarrow$} influence and more balanced contributions across all four groups.}
    \label{fig:arrow_change_gpt2}
\end{figure*}

Figure~\ref{fig:neuron_group_contributions_all_models}b shows that the four groups reduce toxicity projection at different rates when neurons are ranked by their contribution.
\TP{} dominates among the top-ranked neurons, while \AN{} becomes more prominent later, especially in GPT-2-Medium.
Figure~\ref{fig:arrow_change_gpt2} further decodes this trend in GPT-2-Medium, where activation shifts become more evenly distributed in lower-ranked neurons.
% by visualising activation shifts.

Figure~\ref{fig:neuron_group_contributions_all_models}c demonstrates that each group shifts activations according to their orientation relative to the toxic probe, consistent with the pattern observed in Figure~\ref{fig:neuron_group_contributions}c.
%(Figure~\ref{fig:neuron_group_contributions})

% by showing how \textcolor{red}{$\rm TP\downarrow$} and \textcolor{orange}{$\rm AN\downarrow$} reduce toxicity projection at different rates across neuron ranks.

\begin{figure*}[ht]
        \centering
        \includegraphics[width=0.96\textwidth]{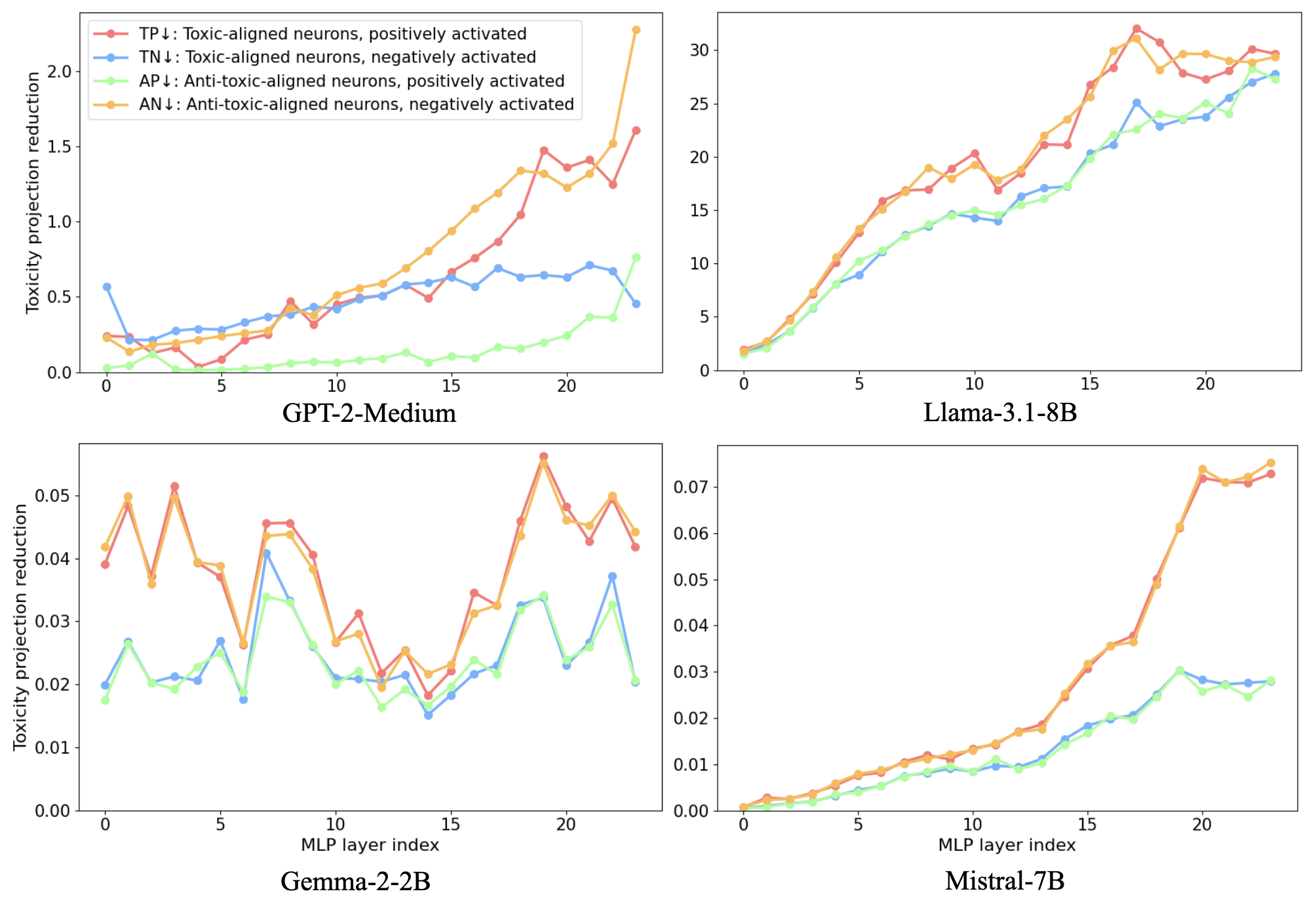}
   \caption{\textit{  
   Layer-wise toxicity projection reduction by neuron group.} Toxicity reduction generally increases across MLP layers under the cumulative group effects, though the upward trend is less evident for Gemma-2-2B.
   % Across models, \TP{} and \AN{} dominate the reduction.
   The upward trend shows that each layer progressively shifts away from toxicity, with the largest toxicity reduction occurring in later layers. }
   \label{fig:per_layer_neuron_groups}
\end{figure*}

Figure~\ref{fig:per_layer_neuron_groups} shows toxicity reduction across layers for all four groups. 
The reduction generally increases through successive MLP layers, reflecting the cumulative effect of activation shifts, though this trend is less pronounced in Gemma-2-2B. These results suggest that layers progressively steer the residual stream away from toxicity, with later layers showing the strongest suppression of toxic outputs. 
The upward trend may be partly due to our use of final-layer probes for extraction.
% In contrast, Gemma-2-2B shows a more uniform level of projection reduction across both early and later layers.

% Table~\ref{tab:merged_broader_groups} shows the proportion of neurons in for broader group across models, as well as the proportion of neurons reducing toxicity projection, which forms our targeted neuron groups (\TP{}, \TN{}, \AP{}, \AN{}). 

\section{More results on activation editing}

In this section, we present more results on activation editing (Section~\ref{subsec:replicating_dpo_with_activation_editing}).

Table~\ref{tab:all_patching_results} extends our probe-based editing results, comparing two selection methods for the top-$\beta$ neurons: descending cosine similarity with probe (main results also in Table~\ref{tab:all_patching_results_main}) and by ascending absolute activations.
While both approaches work, the latter is slightly less effective and fails to surpass DPO for Gemma-2-2B.

As a sanity check, we also patching neurons with increased toxicity projection ($\uparrow$) during DPO and find that they raise toxicity scores across models (Section~\ref{subsec:neuron_groups}).

% Notes: Overall, reduce beta is better to preserve perplexity and F1. But if raise alpha too much at small beta, also break F1 (not because of repetition?)
% Beta=0.5 and 0.3 still reduces F1 with repetition for GPT2 and llama3, but fine for the other two. 
% Llama3 and GPT2 (?) suffers from high perplexity, only resolved when adjust to small beta. 
% The ones that replicate DPO reduces toxicity (not beating steering), but perplexity break for two models and F1 reduces. 

% Lower F1 means more repetitive

\newpage
\begin{table*}[ht]
\renewcommand{\arraystretch}{1.4}  
\setlength{\tabcolsep}{2pt} 
\centering
\small  
\caption{\textit{Toxicity (Toxic), log perplexity (PPL), and F1 scores with activation patching and editing.} 
As a sanity check, patching neurons with increased toxicity projection ($\uparrow$) raises toxicity scores.
In probe-based editing, we compare two samping strategies for the top-$\beta$ neurons: descending cosine similarity with the probe and ascending absolute activation values.
For both approaches, 
\colorbox{green!20}{Green} shows the editing parameters that best compete with DPO while preserving F1 scores.
% and perplexity
% does not reducing toxicity as much as selecting $\beta$ by descending probe alignment
}
\label{tab:all_patching_results}
\begin{tabular}{l|l|p{0.82cm}p{0.73cm}p{0.8cm}|p{0.82cm}p{0.73cm}p{0.8cm}|p{0.82cm}p{0.73cm}p{0.8cm}|p{0.82cm}p{0.73cm}p{0.8cm}}
\toprule
\textbf{Type} & \textbf{Intervention} & \multicolumn{3}{c|}{\textbf{GPT-2-355M}} & \multicolumn{3}{c|}{\textbf{Llama-3.1-8B}} & \multicolumn{3}{c|}{\textbf{Gemma-2-2B}} & \multicolumn{3}{c}{\textbf{Mistral-7B}} \\  
\cline{3-14}
& & \textbf{Toxic} & \textbf{PPL} & \textbf{F1} & \textbf{Toxic} & \textbf{PPL} & \textbf{F1} & \textbf{Toxic} & \textbf{PPL} & \textbf{F1} & \textbf{Toxic} & \textbf{PPL} & \textbf{F1} \\
\hline
\multirow{3}{*}{Baseline} 
& None      & 0.545 & 3.08 & 0.193 & 0.496 & 1.94 & 0.225 & 0.488 & 4.61 & 0.231 & 0.507 & 1.76 & 0.231 \\
% & Forbid toxic tokens & 
% 0.541 & 3.07 & 0.193 &  
% 0.492 & 1.94 & 0.225 & 
% 0.484 & 4.62 & 0.231 & 
% 0.451 & 1.76 & 0.232 \\
& Steering with probe  & 0.310 & 3.19 & 0.191 & 0.335 & 2.72 & 0.187 & 0.260 & 5.52 & 0.228 & 0.350 & 2.23 & 0.220 \\
& DPO       & 0.210 & 3.15 & 0.195 & 0.241 & 2.69 & 0.221 & 0.245 & 5.15 & 0.228 & 0.221 & 2.01 & 0.233 \\
\hline
\multirow{2}{*}{\makecell{Activation\\patching}} &
% Patch \textcolor{red}{$\rm TP\downarrow$} & 
% 0.407 & 3.07 & 0.191 & 
% 0.488 & 1.94 & 0.223 & 
% 0.470 & 5.87 & 0.235 & 
% 0.502 & 1.80 & 0.229 \\
% & Patch \textcolor{red}{$\rm TP\downarrow$}+\textcolor{orange}{$\rm AN\downarrow$} & 
% 0.216 & 3.08 & 0.183 & 
% 0.465 & 1.94 & 0.221 & 
% 0.337 & 4.59 & 0.224 & 
% 0.307 & 1.76 & 0.227 \\
% & Patch \textcolor{red}{$\rm TP\downarrow$}+\textcolor{orange}{$\rm AN\downarrow$}+\textcolor{blue}{$\rm TN\downarrow$} & 
% 0.194 & 3.08 & 0.170 & 
% 0.391 & 1.94 & 0.208 & 
% 0.307 & 4.59 & 0.217 & 
% 0.238 & 1.81 & 0.218 \\
Patch all four groups & 
0.139 & 3.08 & 0.169 & 
0.278 & 1.94 & 0.207 & 
0.260 & 4.58 & 0.213 & 
0.138 & 1.78 & 0.209 \\
& Patch all $\uparrow$ neurons & 
0.853 & 6.05 & 0.154 & 
0.536 & 2.64 & 0.184 & 
0.686 & 4.58 & 0.199 & 
0.611 & 1.78 & 0.199 \\
\hline
\multirow{4}{*}{\makecell{Activation \\editing 
\\
(probe-based,
\\descending
\\cossim)}} 
% by abs cossim (decending)
& {$\alpha=0.01, \beta=0.8$} & 
0.123 & 3.08 & 0.179 & 
0.045 & 2.19 & 0.186 & 
0.199 & 4.54 & 0.188 & 
0.038 & 1.77 & 0.179 \\ 
& {$\alpha=0.01, \beta=0.6$} & 
0.159 & 3.08 & 0.181 & 
0.183 & 2.11 & 0.193 & 
0.200 & 4.56 & 0.201 & 
0.098 & 1.77 & 0.196 \\ 
& {$\mathbf{\alpha=0.01, \beta=0.55}$} & 
\cellcolor{green!20}
0.203 & 3.08 & 0.183 & 
\cellcolor{green!20}
0.241 & 1.96 & 0.196 & 
\cellcolor{green!20}
0.216 & 4.56 & 0.210 & 
\cellcolor{green!20}
0.125 & 1.77 & 0.202 \\ 
& {$\alpha=0.05, \beta=0.5$} & 
0.211 & 3.08 & 0.184 & 
0.299 & 1.96 & 0.200 & 
0.260 & 4.56 & 0.204 & 
0.264 & 1.77 & 0.197\\
\hline
\multirow{4}{*}{\makecell{Activation \\editing
\\
(probe-based, 
\\ascending
\\activation)}} 
% by abs pt activation (ascending)
& {$\alpha=0.01, \beta=0.8$} & 
0.025 & 3.08 & 0.158 &
0.097 & 2.39 & 0.188 & 
0.271 & 4.56 & 0.183 & 
0.154 & 1.77 & 0.196 \\ 
& {$\mathbf{\alpha=0.01, \beta=0.6}$} & 
\cellcolor{green!20}0.075 & 3.07 & 0.178 & 
\cellcolor{green!20}0.204 & 2.26 & 0.198 & 
\cellcolor{green!20}0.295 & 4.57 & 0.202 & 
\cellcolor{green!20}0.218 & 1.77 & 0.201 \\ 
& {$\alpha=0.01, \beta=0.55$} & 
0.111 & 3.08 & 0.175 &
0.258 & 2.25 & 0.203 &
0.330 & 4.57 & 0.199 & 
0.229 & 1.77 & 0.202 \\ 
& {$\alpha=0.05, \beta=0.5$} & 
0.109 & 3.08 & 0.178 & 
0.310 & 1.96 & 0.204 & 
0.331 & 4.58 & 0.204 &
0.251 & 1.77 & 0.193\\ 
% \hline
% \multirow{4}{*}{\makecell{Activation \\editing, 
% \\(probe-free,
% \\descending
% \\cossim)}} 
% % by descending cossim
% & {$\alpha=0.01, \beta=0.8$} & 
% 0.139 & 3.08 & 0.176 &
% 0.116 & 5.82 & 0.200 &
% 0.218 & 4.54 & 0.180 &
% 0.057 & 1.77 & 0.191\\ 
% & {$\mathbf{\alpha=0.01, \beta=0.6}$} &
% \cellcolor{green!20}0.238 & 3.08 & 0.178&
% \cellcolor{green!20}0.258 & 2.28 & 0.210 &
% \cellcolor{green!20}0.216 & 4.57 & 0.203 &
% \cellcolor{green!20}0.162 & 1.77 & 0.200 \\ 
% & {$\alpha=0.01, \beta=0.55$} & 
% 0.282 & 3.08 & 0.180 &
% 0.318 & 2.24 & 0.204 &
% 0.250 & 4.58 & 0.198 &
% 0.239 & 1.77 & 0.201 \\ 
% & {$\alpha=0.05, \beta=0.5$} & 
% 0.275 & 3.08 & 0.180&
% 0.322 & 2.21 & 0.201 &
% 0.271 & 4.58 & 0.193&
% 0.244 & 1.77 & 0.199 \\ 
% \hline
% \multirow{4}{*}{\makecell{Activation \\editing
% \\
% (toxic embed,
% \\descending
% \\cossim)}} 
% % by abs pt activation (ascending)
% & {$\alpha=0.01, \beta=0.8$} & 
% & & &
% 0.116 & \textcolor{red}{5.82} & 0.200 &
% & & &
% 0.057 & 1.77 & 0.191\\ 
% & {$\alpha=0.01, \beta=0.6$} &
% & & &
% 0.258 & 2.28 & 0.210 &
% & & &
% 0.162 & 1.77 & 0.201 \\ 
% & {$\alpha=0.01, \beta=0.55$} & 
% 0.282 & 3.08 & 0.180 &
% 0.318 & 2.24 & 0.204 &
% 0.250 & 4.58 & 0.200 &
% 0.239 & 1.77 & 0.200 \\ 
% & {$\alpha=0.05, \beta=0.5$} & 
% & & &
% & & &
% & & &
% & & \\ 
\bottomrule
\end{tabular}
\end{table*}

\end{document}